\pdfoutput=1

\documentclass[11pt,letter]{article}
\usepackage[papersize={8.5in,11in},margin=1in]{geometry}

\usepackage[format=hang,font={footnotesize}]{caption}
\usepackage[numbers,sort,compress]{natbib} 
\usepackage{graphicx}

\usepackage[title]{appendix}
\usepackage{amsthm}
\usepackage{amsmath}
\usepackage{amssymb}
\usepackage{mathtools}
\usepackage{url}
\usepackage{subfigure}
\usepackage{wrapfig}
\usepackage{color}
\usepackage{enumitem}
\usepackage{grffile}
\usepackage{marvosym}
\usepackage{complexity}
\usepackage{setspace}
\usepackage{multicol}
\usepackage{multirow}
\usepackage{array}
\usepackage{booktabs}
\usepackage{lettrine,Zallman}

\usepackage{calligra}

\usepackage{tcolorbox}
\usepackage{url}
\usepackage{color}
\usepackage{tikz}
\usetikzlibrary{positioning,shapes,arrows}
\usepackage{algorithm,algorithmicx} \usepackage[noend]{algpseudocode}

\usepackage{xspace}
\usepackage{dsfont} \usepackage{pifont} \usepackage{musicography}
\usepackage{bbm}
\usepackage{comment}
\usepackage{rotating}
\usepackage{stmaryrd}

\usepackage{tikz}
\usetikzlibrary{shapes,arrows.meta}
\usetikzlibrary{arrows}

\graphicspath{{figs/}}

\newcommand{\dtlinkcolor}{{0.8 0.8 1}} \usepackage[hyperindex=true,pdfpagemode=UseOutlines,bookmarksnumbered=true,bookmarksopen=true,bookmarksopenlevel=2,pdfstartview=FitH,pdfborder={0 0 1},linkbordercolor=\dtlinkcolor,citebordercolor=\dtlinkcolor,urlbordercolor=\dtlinkcolor,pagebordercolor=\dtlinkcolor]{hyperref}
\hypersetup{pageanchor=false,pdfpagelabels}
\usepackage{hyperxmp} \hypersetup{keeppdfinfo,pdfmetalang={en}} \hypersetup{pdftitle={Physics of Language Models: Part 4.2, Canon Layers at Scale where Synthetic Pretraining Resonates in Reality},pdfauthor={Zeyuan Allen-Zhu},pdfsubject={Large-scale controlled pretraining and language model architecture design},pdfkeywords={Physics of Language Models, Canon layers, architecture design, controlled pretraining, linear attention, state space models, Mamba2, GLA, GatedDeltaNet, Transformer, reproducibility}}
\usepackage[align=center,shadow=true,shadowsize=5pt,nobreak=true,framemethod=tikz,style=0,skipabove=2pt,skipbelow=1pt,innertopmargin=-3pt,innerbottommargin=3pt,innerleftmargin=5pt,innerrightmargin=5pt,leftmargin=-2pt,rightmargin=-2pt]{mdframed}
\usetikzlibrary{shadows}

\allowdisplaybreaks[1]
\theoremstyle{plain} \setitemize{itemsep=0mm, topsep=1mm, leftmargin=8mm}
\setenumerate{itemsep=0mm, topsep=1mm, leftmargin=8mm}
\AtBeginDocument{%
 \abovedisplayskip=6pt minus 1pt
 \abovedisplayshortskip=4pt plus 1pt
 \belowdisplayskip=6pt minus 1pt
 \belowdisplayshortskip=4pt plus 1pt
}

\newcommand{\parhead}[1]{\smallskip \noindent {\bfseries\boldmath\ignorespaces #1.}\hskip 0.9em plus 0.3em minus 0.3em}
\newcommand{\parheadno}[1]{\smallskip \noindent {\bfseries\boldmath\ignorespaces #1}\hskip 0.9em plus 0.3em minus 0.3em}

\newtheorem{innercustomres}{Result}
\newenvironment{sresult}[1]
  {\renewcommand\theinnercustomres{#1}\innercustomres}
  {\endinnercustomres}

\newtheorem*{theorem*}{Theorem}
\newtheorem{theorem}{Theorem}[section]

\newtheorem*{assumption*}{Assumption}
\newtheorem*{question*}{Question}

\newtheorem*{rep@theorem}{\rep@title}
\newcommand{\newreptheorem}[2]{%
\newenvironment{rep#1}[1]{%
 \def\rep@title{#2 \ref{##1}}%
 \begin{rep@theorem}}%
 {\end{rep@theorem}}}

\newreptheorem{theorem}{Theorem}
\newreptheorem{lemma}{Lemma}
\newreptheorem{proposition}{Proposition}
\newreptheorem{claim}{Claim}
\newreptheorem{definition}{Definition}
\newreptheorem{assumption}{Assumption}

\theoremstyle{definition}

\theoremstyle{remark}\newtheorem{remark}[theorem]{Remark}
\newtheorem*{remark*}{Remark}

\numberwithin{equation}{section}

\makeatletter
\let\c@table\c@figure
\makeatother

\newcommand{\namedref}[2]{\mbox{\hyperref[#2]{#1~\ref*{#2}}}}

\newcommand{\sectionref}[1]{\namedref{Section}{#1}}
\newcommand{\appendixref}[1]{\namedref{Appendix}{#1}}

\newcommand{\resultref}[1]{\namedref{Result}{#1}}

\newcommand{\figureref}[1]{\namedref{Figure}{#1}}
\newcommand{\figurerefb}[2]{\mbox{\hyperref[#1]{Figure~\ref*{#1}#2}}}
\newcommand{\tableref}[1]{\namedref{Table}{#1}}

\newcommand{\footnoteref}[1]{\namedref{Footnote}{#1}}
\newcommand{\equationref}[1]{\mbox{\hyperref[#1]{(\ref*{#1})}}}
\renewcommand{\eqref}{\equationref}

\makeatletter
\newcommand\xLongLeftRightArrow[2][]%
  {\ext@arrow 0099{\LongLeftRightArrowfill@}{#1}{#2}}
\def\LongLeftRightArrowfill@
  {\arrowfill@\Leftarrow\Relbar\Rightarrow}
\newcommand\xLongRightArrow[2][]%
  {\ext@arrow 0099{\LongRightArrowfill@}{#1}{#2}}
\def\LongRightArrowfill@
  {\arrowfill@\Relbar\Relbar\Rightarrow}
\makeatother

\newcommand{\defem}[1]{\textsf{\emph{#1}}}
\newcommand{\arxivbf}[1]{\textbf{#1}}
\newcommand{\arxivbblue}[1]{\bblue{#1}}
\newcommand{\arxivem}{\expandafter\emph}
\newcommand{\nipswidth}[2]{ #2} \newcommand{\arxivbreak}{\\} \newcommand{\detext}[1]{\textsf{#1}}

\renewcommand{\hbar}{\breve{h}}

\usetikzlibrary{tikzmark}
\usetikzlibrary{calc}

\errorcontextlines\maxdimen

\newcommand{\ALGtikzmarkcolor}{black}\newcommand{\ALGtikzmarkextraindent}{4pt}\newcommand{\ALGtikzmarkverticaloffsetstart}{-.5ex}\newcommand{\ALGtikzmarkverticaloffsetend}{-.5ex}\makeatletter
\newcounter{ALG@tikzmark@tempcnta}

\newcommand\ALG@tikzmark@start{%
    \global\let\ALG@tikzmark@last\ALG@tikzmark@starttext%
    \expandafter\edef\csname ALG@tikzmark@\theALG@nested\endcsname{\theALG@tikzmark@tempcnta}%
    \tikzmark{ALG@tikzmark@start@\csname ALG@tikzmark@\theALG@nested\endcsname}%
    \addtocounter{ALG@tikzmark@tempcnta}{1}%
}

\def\ALG@tikzmark@starttext{start}
\newcommand\ALG@tikzmark@end{%
    \ifx\ALG@tikzmark@last\ALG@tikzmark@starttext
                    \else
        \tikzmark{ALG@tikzmark@end@\csname ALG@tikzmark@\theALG@nested\endcsname}%
        \tikz[overlay,remember picture] \draw[\ALGtikzmarkcolor] let \p{S}=($(pic cs:ALG@tikzmark@start@\csname ALG@tikzmark@\theALG@nested\endcsname)+(\ALGtikzmarkextraindent,\ALGtikzmarkverticaloffsetstart)$), \p{E}=($(pic cs:ALG@tikzmark@end@\csname ALG@tikzmark@\theALG@nested\endcsname)+(\ALGtikzmarkextraindent,\ALGtikzmarkverticaloffsetend)$) in (\x{S},\y{S})--(\x{S},\y{E});%
    \fi
    \gdef\ALG@tikzmark@last{end}%
}

\apptocmd{\ALG@beginblock}{\ALG@tikzmark@start}{}{\errmessage{failed to patch}}
\pretocmd{\ALG@endblock}{\ALG@tikzmark@end}{}{\errmessage{failed to patch}}
\makeatother

\definecolor{mygreen}{RGB}{80,180,0}
\definecolor{figorange}{RGB}{235,104,52} \definecolor{figgreen}{RGB}{27,175,122} \definecolor{figblue}{RGB}{42,120,214}

\newcommand{\bblue}[1]{{\color{blue}\textbf{#1}}}

\usepackage{lineno}

\usepackage{varwidth}
\usepackage{CJKutf8}

\begin{document}

\newlength{\imgwidthBase}

\title{Physics of Language Models: Part 4.2, \\ Canon Layers at Scale \\ where Synthetic Pretraining Resonates in Reality}

\date{December 9, 2025\\
\medskip
\normalsize (version 2.0)%
\thanks{\textbf{On the date above.} By that date, recipes, code, and results for all 64 models were public at \href{https://github.com/facebookresearch/PhysicsLM4}{\texttt{github.com/facebookresearch/PhysicsLM4}}. The 16 Transformer checkpoints were released on July 23, 2025; December 9 added linear recipes/results. The findings were presented on January 12, 2026 (\href{https://youtu.be/niHNWJxmkW0}{\texttt{youtu.be/niHNWJxmkW0}}). Only this manuscript's writing was delayed. \textbf{V2} polishes the writing.
\newline
\indent
\textbf{Future updates:} on \href{https://physics.allen-zhu.com/part-4-architecture-design/part-4-2}{\texttt{physics.allen-zhu.com}}.
\newline
\indent
We thank FAIR at Meta, where all experiments presented in this paper were conducted and their results were open-sourced. We also thank Xiaoli Xu for many helpful conversations while planning and preparing this manuscript.
}
}

\newcommand{\authorname}[1]{\makebox[6cm][c]{#1}}
\author{
\authorname{Zeyuan Allen-Zhu} \\
\texttt{\href{mailto:zeyuan2026@allen-zhu.com}{\color{black}zeyuan2026@allen-zhu.com}} \\
Allen-Zhu Research
}

\maketitle

\begin{abstract}
Do standard benchmarks see the architectural differences that matter? We study this question with \arxivem{64 from-scratch pretraining runs} of Transformers and three linear-model families at 1.3--8B parameters on 1--2T Nemotron-CC tokens.

In academic-scale pretraining (1.3B parameters, 100B tokens), \arxivem{architectural differences can all but disappear} when the models are carefully tuned. Instead, at 1--2T tokens they become measurable: Canon layers~\cite{Allen2025-canon}, which add local token mixing, consistently improve Transformers, \emph{saving about 30\% of pretraining tokens} for the same MMLU accuracy at 8B. We also introduce \arxivem{GLA5 and GDN2},  which improve gated linear attention and Gated DeltaNet with low-rank output gates.   Surprisingly, with  Canon, the simplest linear model, gated linear attention, \arxivem{matches Gated DeltaNet and outperforms  Mamba2} even at 8B/1T.  Therefore, the extra machinery of Gated DeltaNet and Mamba2 may do little more than mix neighboring tokens. 

\textbf{Even at 8B/1T}, however, high scores on standard benchmarks can hide weaknesses in basic in-context skills. In particular, linear models that match or beat Transformers on these benchmarks \emph{cannot, in fact,\ reliably retrieve information from a short 500-token context}. Moreover, \emph{no base model, Transformer or linear, reliably beats random guessing} on arguably the simplest latent two-hop question (without chain-of-thought prompting), stated in a few short sentences, \textbf{even at 8B/1T}. Larger models may well learn this skill, but our 8B runs, at roughly \$100k each, still cannot reliably rank architectures on it, and deeper latent reasoning would presumably need even larger runs for every design.

These findings \textbf{\empty challenge  traditional architecture research} based primarily on standard benchmark scores. \arxivem{In contrast}, small synthetic playgrounds, which  predicted the architectural differences above at $<$100M parameters,  can reveal what these benchmarks miss. For practitioners,  one lesson is to include \arxivem{in-context reasoning} examples in pretraining data from an early stage, rather than reserve them  for later mid-training, let alone post-training.

\end{abstract}

\thispagestyle{empty}
\clearpage
\setcounter{page}{1}

\section{Introduction}
\label{sec:intro}

\lettrine[lines=3]{A}{}rchitecture
research usually compares language-model designs by their pretraining loss or by their scores on standard benchmarks. Such comparisons are meaningful only with carefully tuned baselines and matched model budgets. Yet when a carefully controlled comparison produces similar scores, the conclusion is not obvious: are the architectures equally effective, or is the experiment simply unable to tell them apart? The answer decides which designs the field adopts and which promising ideas it discards.

Part 4.1~\cite{Allen2025-canon} ran into exactly this question in \emph{academic-scale real-life pretraining} (1.3B parameters, 100B tokens). It tuned every implementation carefully --- its gated linear attention (GLA) and Gated DeltaNet (GDN) scores exceeded those reported in the original studies --- and controlled model budgets and training conditions. Under these controlled comparisons, most architectural differences disappeared: standard knowledge and commonsense scores clustered closely across Transformers and most linear-model variants~\cite[Result 12]{Allen2025-canon}.\footnote{Clearly weaker variants, such as GLA without a short convolution, remained distinguishable.} Even the gains from adding \emph{Canon layers}, lightweight direct connections between neighboring tokens, were generally not clearly visible.

Part 4.1's answer to this problem is a \emph{synthetic pretraining playground}, in which controlled training examples make reasoning, knowledge, and hierarchical-structure capabilities measurable below 100M parameters. The playground reveals differences that academic-scale pretraining cannot, and it was the playground that motivated Canon layers in the first place. This leaves a natural question: do architectural insights from a small playground hold in much larger real-life pretraining?

 We answer this question  with \arxivbf{64 from-scratch pretraining runs} of Transformers (Llama), Mamba2,  GLA, and GDN, at 1.3--8B parameters on \mbox{1--2T} high-quality Nemotron-CC tokens.  We highlight two designs that came from the synthetic playground: Canon layers, and our \emph{GLA5} and \emph{GDN2}, which improve GLA and GDN with low-rank output gates.  To keep the comparisons fair, we tune every model's learning rate, strictly align the linear models' parameter allocation and recurrent memory, and compare full training curves  rather than  final checkpoints. The resulting baselines are strong: on the ten benchmarks that Nvidia also reports, our vanilla 8B Llama scores 2.5 points above Nvidia's Transformer baseline trained on the same data at the same budget\empty , and LlamaCanon, which adds Canon layers to every Llama sublayer, matches Llama-3.1-8B using 1/15 of its training tokens.  

On one hand, our results show that the synthetic playground's predictions from Part 4.1 largely hold at scale:
  Canon consistently improves Transformers, and with Canon, simple gated linear attention matches Gated DeltaNet and outperforms Mamba2, at least up to 8B/1T, with both GLA and GDN in our improved, strictly aligned forms (\namedref{Results}{res:3}--\ref{res:5}).  On the other hand, the same  runs  show that high standard-benchmark scores do not guarantee basic in-context skills.\ In particular, linear models that match or beat Transformers on standard benchmarks cannot, in fact, reliably retrieve information from a short 500-token context. Moreover, no base model, Transformer or linear, reliably beats random guessing on arguably the simplest latent two-hop question (without chain-of-thought prompting), stated in a few short sentences, even at 8B/1T.

\subsection{Seven Findings}
\label{sec:intro:roadmap}

\begin{itemize}
\item \textbf{Canon consistently improves Transformers (\resultref{res:1}).} In every tested setting (1.3--8B, 1--2T tokens), even the worst LlamaCanon run beats the best vanilla run on \texttt{lm-eval}, our 13-benchmark average --- a gain that academic-scale pretraining (1.3B/100B) could not reveal.
\item \textbf{Canon saves about 30\% of pretraining tokens at 8B (\resultref{res:2}).} That is, LlamaCanon first reaches vanilla Llama's best final MMLU accuracy with about 30\% fewer tokens, as read from full training curves, and this saving shows no sign of shrinking with scale.
\item  \textbf{Low-rank output gates give stronger linear-model baselines (\resultref{res:3}).} In GLA and GDN, a full-rank output gate can take 1/3 of the recurrent layer's parameters. Our GLA5 and GDN2 improve both by using a low-rank gate instead and giving the saved parameters to the rest of the layer. All three linear models are also strictly aligned: they split parameters the same way and use equally large recurrent states. \item \textbf{With Canon, GLA5 matches GDN2 (\resultref{res:4}).} Because GLA5 and GDN2 are strictly aligned, comparing them largely isolates how they update their recurrent memory: GLA5 uses plain gated accumulation, whereas GDN2 adds the delta rule. GDN2 leads GLA5 at every size without Canon, but with Canon they come within 0.3 points on \texttt{lm-eval} and their training losses nearly coincide.
\item \textbf{\empty Horizontal mixing dominates architectural sophistication  (\resultref{res:5}).}  Canon helps most where an architecture's own local mixing is weakest:  Transformers and GLA5 gain most overall, GDN2 gains about half as much as GLA5, and Mamba2 gains little. As a result, across 1.3--8B, GLA5 rises from behind GDN2 to tie it for first on \texttt{lm-eval} among the three strictly aligned linear models.
\item \textbf{Standard scores hide a retrieval weakness (\resultref{res:6}).} Linear models that match or beat Transformers on standard benchmarks are  in fact  unreliable at retrieving information from their input. To expose this, we ask each model for a birth year from a few short statements interleaved with $L$ unrelated tokens. Already at $L=500$, every linear model scores below the Transformers of its size.
\item \textbf{Latent two-hop reasoning barely emerges at 8B/1T (\resultref{res:7}).} Given three birth years such as ``Ann was born in 1987'' and three statements such as ``Bob was born in the same year as Ann'', the model must complete ``Bob was born in the year of'' (without chain-of-thought prompting).  This is arguably the simplest two-hop question, yet no model reliably beats random guessing, even without any background text. %
\footnote{With three candidate years, random guessing scores 33.3\%. The 8B Transformers are the only partial exception, at about 50\% without background text, but this advantage mostly vanishes with 500 background tokens (\resultref{res:7}).}\end{itemize}

\subsection{Four Lessons We Did Not Expect}
\label{sec:intro:resonance}

Several of these results confirm what the synthetic playground predicted (\tableref{tab:resonance}), whereas others, such as the 30\% token saving, the low-rank output gates, and the near-random latent two-hop accuracy at 8B/1T, are new to this study. Together, they reveal four lessons we did not expect.

\parhead{Real-life pretraining misses differences that the synthetic playground reveals}
We set out to test the playground against real-life pretraining, treating the latter as ground truth. For telling architectures apart, however, the roles turned out closer to the reverse. Not only can academic-scale real-life pretraining (1.3B/100B) hide most architectural differences, but even 8B/1T pretraining leaves models near random guessing on latent two-hop reasoning (\resultref{res:7}).  Comparisons of architectures or optimizers at this scale or below thus become highly questionable for such reasoning skills.  Training larger models would not remove this problem: a run large enough to learn latent two-hop reasoning would presumably still fail at latent three-hop reasoning, and such runs are too costly to repeat for every design being compared. The synthetic playground, by contrast, already revealed such architectural differences, including in multi-hop reasoning, at $\leq$100M parameters and \mbox{$100$--$1000\times$} lower cost per run (\sectionref{sec:res7:lessons}).

\parhead{Mixing neighboring tokens can matter more than a more elaborate recurrent design}
GLA simply accumulates information in its recurrent memory, whereas GDN adds a delta rule that corrects what is already stored, and Mamba2 adds a selective state-space (SSM) update. Without Canon, this extra machinery seems to pay off: GDN2 leads GLA5 at every size. Once all three are equipped with Canon, however, GLA5 matches GDN2 and outperforms Mamba2 at every size, at least up to 8B/1T. The extra machinery may therefore mainly supply short-range mixing, which Canon provides directly at under 0.2\% of the parameters. This reflects a central question of deep learning, one that also guides the synthetic playground: what a ``clever'' mechanism can express matters less than what training actually learns --- \emph{existence is cheap; learnability is everything}. We return to this principle in Results~\ref{res:3} and~\ref{res:5} (Sections~\ref{sec:linear-upgrade:gate} and~\ref{sec:res5}).

\parhead{A benchmark win can conceal a basic deficit}
Linear-model papers typically report matching or beating Transformers on standard \texttt{lm-eval} benchmarks~\cite{yang2023gated,dao2024transformers,yang2024gated}. Our runs at 1.3--8B on 1T tokens not only confirm this, but show that with Canon, even GLA5 beats both Llama and LlamaCanon at every size. However, \textbf{this win can be highly misleading}. Linear models are good at storing knowledge in their weights, which is much of what these benchmarks test, but even at 8B/1T they learn to recall information from their input far more slowly than Transformers and become unreliable with just 500 tokens of context (\resultref{res:6}). They thus cannot yet replace attention even over a few thousand tokens, let alone over the very long contexts they are often expected to handle. This may be the real reason hybrid models~\cite{lieber2024jamba,Qwen3next,tiifalconh1} keep attention layers. Standard scores alone are therefore an incomplete basis for architecture design.

\parhead{Include reasoning data earlier}
Even after 1T tokens of filtered and knowledge-augmented Nemotron-CC data, no model reliably beats random guessing on arguably the simplest latent two-hop question. Post-training tends to sharpen skills that a model already has rather than add missing ones, so such skills can be hard to teach once pretraining is over (\sectionref{sec:res6:sft}). For practitioners, we therefore suggest including in-context reasoning examples (\emph{real or synthetically generated}) in pretraining data from an early stage, rather than reserving them for mid-training, let alone post-training. Even simple two-hop examples should appear well before 1T pretraining tokens, so that the model's internal states start preparing for reasoning while it acquires knowledge, rather than being reshaped afterwards.

\section{Experimental Setup}
\label{sec:setup}

We train \textbf{64 models from scratch} at three sizes. Within each size, every architecture shares the same shape: 1.3B (24 layers, 2048 dim), 3B (26 layers, 3072 dim), and 8B (32 layers, 4096 dim), where each layer consists of an attention or recurrent module followed by an MLP. Of these runs, 16 compare Llama with LlamaCanon on 1T tokens (plus 1.3B on 2T tokens), while 48 compare Mamba2, GLA5, and GDN2 (\empty our improved GLA and GDN), with and without Canon, on 1T tokens. Scale matters because in academic-scale pretraining (1.3B/100B), architectural differences can all but disappear when the models are carefully tuned~\cite[Result 12]{Allen2025-canon}. But a \emph{larger experiment} is \emph{\textbf{not}} automatically a more \emph{trustworthy} one: inherited baselines can be undertuned, published architectures can differ in parameter allocation or memory rather than only in design, and a single run can land on a lucky checkpoint. We therefore control four aspects of the experiment before comparing architectures. \appendixref{app:config} gives the full configurations and the reasoning behind each choice.\looseness=-1

\parhead{1. Fix a strong data mix}
All runs use the same high-quality split of Nemotron-CC v1~\cite{su2025nemotroncc}, which combines quality filtering with LLM rewriting, an application of the \emph{knowledge augmentation} introduced in Part 3.1~\cite{AL2023-knowledge}. By contrast, academic-scale studies before this work, including Part 4.1, typically trained on corpora such as FineWeb-Edu~\cite{penedo2024fineweb}, SlimPajama~\cite{cerebras2023slimpajama}, or DCLM~\cite{li2024datacomp}, which are heavily filtered but not rephrased. Part 3.1 showed that without such rephrasing, knowledge learning can be far less efficient. The gap is large: Nvidia reports that in same-budget 8B/1T runs, models trained on Nemotron-CC score 3 and 7 points higher than those trained on DCLM and FineWeb-Edu on the 10-task benchmark average of \figureref{fig:scatter}~\cite{su2025nemotroncc}. Stronger data thus lets models learn more skills within the same budget, making architectural comparisons more meaningful.

\parhead{2. Tune every model, not just the new one}
Hyperparameters transferred from one method to another can disadvantage the baseline~\cite{wen2025fantastic}, so we train every configuration at its 2--3 best learning rates.%
\footnote{\label{fn:best-lr}We first tuned the learning rate over a much wider range; the 2--3 rates we train with are the best ones from that search. The same holds for every learning-rate grid in this paper.}
As a result, our vanilla 8B Llama scores 2.5 points above the score Nvidia reports for its Transformer baseline trained on the same data and budget~\cite{su2025nemotroncc} (\appendixref{sec:setup:lr}). Since an 8B/1T run costs about $4\times10^4$ H200 GPU-hours (roughly \$100k), we sweep learning rates rather than repeat seeds: the sweep also shows whether a conclusion survives a second run, and whether the best rate lies inside the grid. Summary tables report each task's best score across learning rates; \appendixref{app:all-lr} reports every run separately.

\parhead{3. Align resources and training}
Each Canon pair compares the same model with and without Canon, and everything except the Canon recipe is aligned. Within each size and token budget, the training tokens, data order, seed, tokenizer, depth, hidden width, implementation, and evaluation pipeline are shared, and the learning-rate grids largely overlap.%
\footnote{Transformers train with context length 4096, whereas the linear models use 16384, 12288, and 4096 at 1.3B, 3B, and 8B ($4\times$, $3\times$, and $1\times$; \tableref{tab:config}), because a linear model's cost per token does not grow with context length while attention's does. Both train on the same number of tokens, but the linear models take 2--3$\times$ as many optimizer steps with 2--3$\times$ smaller batches. At 1.3B and 3B, the linear models are trained at their three best learning rates instead of two (\footnoteref{fn:best-lr}). These differences favor the linear models, which makes many of our main findings, such as their retrieval deficit, more robust.}
The linear models GLA5, GDN2, and Mamba2 are \emph{additionally} strictly aligned on almost everything possible: parameter allocation, recurrent memory, heads, context length, batch size, and training steps, so that their recurrent memory designs are compared under the same resource constraints (\resultref{res:3}). In particular, they even train on identical data batches.

\parhead{4. Measure the base model}
We evaluate the base models before SFT, RL, or reasoning-specific mid-training. Otherwise, a difference could come from post-training objectives or data rather than from the architecture, not to mention that post-training would add more noise to already noisy real-life comparisons. The base models show what pretraining itself has established, and \sectionref{sec:res6:sft} explains why a skill missing at this point can be hard to add during post-training.

\parhead{Evaluation}
Each final checkpoint is evaluated with the standard harness~\cite{eval-harness} (\appendixref{app:eval}) on \texttt{lm-eval}, the average of 13 knowledge and commonsense benchmarks such as MMLU, HellaSwag, and ARC; WikiText and LAMBADA perplexity; Just Read Twice (JRT; 12 metrics, e.g., SQuAD and TriviaQA)~\cite{arora2024just}; BabiLong (20 metrics)~\cite{kuratov2024babilong}; NIAH-S, a single-needle retrieval test~\cite{hsieh2024ruler}; and two of \emph{our own} controlled tasks, 1-hop-$L$ and 2-hop-$L$, for one- and two-hop retrieval (\sectionref{sec:beyond}). Training curves come from our Lingua-based codebase~\cite{meta2024lingua}, which evaluates every \texttt{lm-eval} benchmark, plus TriviaQA, every few thousand steps. All recipes, code, results, and curves are public, as are the 16 Transformer checkpoints (\appendixref{app:repro}).

\subsection{Canon Layers, Briefly}
\label{sec:canon-recap}

A Transformer moves information between tokens only through attention, even when a token needs information only from the few tokens right before it --- using global attention for such a local transfer is like shooting a bird with a cannon. Short causal convolutions that supply such local mixing are not new: Mamba, GLA, and GDN apply one after their input projections, following H3~\cite{fu2022hungry}, and Primer~\cite{so2109primer} applies one to queries, keys, and values (\sectionref{sec:related}). Part 4.1~\cite{Allen2025-canon} found that this cheap \emph{horizontal mixing} is among the most important components of an architecture, worked out how best to add it, and called the resulting layers \emph{Canon layers}. For hidden states $h_t\in\mathbb{R}^m$, each coordinate receives, as a residual, a learned weighted sum of its values at the current and three preceding tokens:
\begin{equation}\label{eqn:canon}
h'_t = h_t + w_0 \odot h_t + w_1 \odot h_{t-1} + w_2 \odot h_{t-2} + w_3 \odot h_{t-3},
\qquad w_0,\ldots,w_3 \in \mathbb{R}^m .
\end{equation}
Part 4.1's recipe implements this sum as a residual \texttt{conv1d}, with a larger initialization than linear models use for their built-in convolution, and inserts it at up to four positions in each layer: A, before the attention or recurrent module; B, after that module's input projections; C, before the MLP; and D, after both MLP projections, before applying SiLU to one branch and taking the elementwise product. Prior designs rarely use positions A, C, and D.\footnote{Since Part 4.1's proof of concept{} and the public release of this work's code and results (2025) and talk (January 2026), two open mixture-of-experts models far larger than ours have added short convolutions: Inkling~\cite{thinkingmachines2026inkling} (975B) and Marin 535B-A23B~\cite{dial2026marin}. Both use position B (on keys and values in Inkling, on keys in Marin) and an output-side variant of A and C, applied to the attention and MLP outputs before they rejoin the residual stream.} Canon at all four positions, denoted Canon-ABCD, adds only 0.07--0.15\% to a Transformer's parameters. The convolution is only one way to implement Canon: in Part 4.1, even fixed random weights gave large gains. We use \texttt{conv1d} because short causal convolutions are already widely supported: besides training kernels, inference engines such as vLLM~\cite{kwon2023efficient} implement them for Mamba-style models, so Canon can reuse these kernels at deployment. As for training speed, Part 4.1 reported step-time overheads of 12.4\% (forward) and 14.1\% (backward) for Canon-ABCD, but with a naive implementation that relied on \texttt{torch.compile} defaults. We have since fused each Canon layer with a neighboring operation in a preliminary, not yet fully optimized implementation, and compared it with the best kernels of Transformer Engine~2.20.2~\cite{nvidia2026transformerengine} in Megatron~\cite{shoeybi2019megatron}: Canon-B, fused into the QKV split and RoPE, makes a training step 0.8--2.0\% \emph{faster}; Canon-D, fused with SwiGLU, costs 0.7--0.9\%; and Canon-A and Canon-C, fused with RMSNorm, together cost 3.2--3.6\%. Altogether, Canon-ABCD adds only 2.1--3.3\% to the step time (\appendixref{app:canon-cost}).

The linear models already include a non-residual short convolution at position B, denoted \mbox{Canon-b}. We keep it and add Canon at A, C, and D (\mbox{Canon-AbCD}). Their gains are thus measured against a baseline that already mixes nearby tokens --- a deliberately conservative comparison.

For Transformers, \bblue{LlamaCanon} combines Canon-ABCD with rotary position encoding (RoPE) on only 1/4 of each head's dimensions. Part 4.1 found that Canon takes over part of the nearby-token work that positional encoding otherwise does, which motivates this choice. All Transformer comparisons evaluate this combined recipe. \appendixref{app:canon-details} gives the construction, \empty a schematic (\figureref{fig:canon-illustrate}),  exact parameter counts, and the RoPE comparisons.

\section{Transformers: Gains That Academic Scale Hides}
\label{sec:transformer-results}

\subsection{Result 1: Canon Consistently Improves Transformers at 1.3--8B}
\label{sec:res1}

We compare Llama and LlamaCanon in four settings: 1.3B parameters with 1T or 2T tokens, and 3B or 8B parameters with 1T tokens. Training each model at its two best learning rates (\footnoteref{fn:best-lr}) gives 16 runs and eight matched pairs. Within a pair, the only change is the Canon recipe, including its quarter RoPE; data, data order, seed, and schedule are identical.

\begin{mdframed}
\begin{sresult}{1}[\tableref{tab:trans-main}]
\label{res:1}
In all four Transformer settings, \defem{even the worst LlamaCanon run beats the best vanilla Llama run} on \texttt{lm-eval}. When each model is scored at its best learning rate for each task, Canon adds 0.8--1.5 points on \texttt{lm-eval} and 1.3--2.9 on MMLU, gains that seed noise hid at academic scale (1.3B/100B).
\end{sresult}
\end{mdframed}

\begin{table}[ht!]
\centering
\small\setlength{\tabcolsep}{4.4pt}
\renewcommand{\arraystretch}{1.16}\begin{tabular}{@{}llcccccc@{}}
\toprule
& & \multicolumn{4}{c}{\textbf{Accuracy} $\uparrow$} & \multicolumn{2}{c}{\textbf{Perplexity} $\downarrow$} \\
\cmidrule(lr){3-6}\cmidrule(l){7-8}
\textbf{size / tokens} & \textbf{model} & \texttt{lm-eval}(13) & \textbf{MMLU} & \textbf{JRT}(12) & \textbf{BabiLong}(20) & \textbf{Wiki ppl} & \textbf{LMB ppl} \\
\midrule
\multirow{2}{*}{8B / 1T}
 & Llama       & 66.9 & 60.6 & 64.7 & 51.3 & 9.60 & 4.60 \\
 & LlamaCanon  & \textbf{67.7} & \textbf{62.4} & \textbf{64.8} & \textbf{55.3} & \textbf{9.19} & \textbf{4.20} \\
\midrule
\multirow{2}{*}{3B / 1T}
 & Llama       & 62.3 & 53.2 & 58.1 & 42.4 & 11.45 & 5.47 \\
 & LlamaCanon  & \textbf{63.8} & \textbf{56.1} & \textbf{59.8} & \textbf{49.4} & \textbf{11.09} & \textbf{5.09} \\
\midrule
\multirow{2}{*}{1.3B / 1T}
 & Llama       & 57.7 & 44.5 & 56.6 & 36.0 & 13.67 & 7.54 \\
 & LlamaCanon  & \textbf{58.9} & \textbf{47.0} & \textbf{57.8} & \textbf{45.3} & \textbf{13.17} & \textbf{7.07} \\
\midrule
\multirow{2}{*}{1.3B / 2T}
 & Llama       & 57.7 & 45.5 & 56.4 & 34.1 & 13.30 & 7.03 \\
 & LlamaCanon  & \textbf{58.7} & \textbf{46.8} & \textbf{57.7} & \textbf{43.7} & \textbf{12.86} & \textbf{6.86} \\
\bottomrule
\end{tabular}
\caption{\label{tab:trans-main}\textbf{LlamaCanon improves all six summary metrics in each of the four Transformer settings.} The number in parentheses is how many accuracy metrics a column averages, and Wiki and LMB denote WikiText and LAMBADA perplexity. Each task takes the better of two learning rates before averaging, so a row can combine runs. Bold marks the better model in each setting. Per-task scores are in \figureref{fig:trans-table}, and every individual run in \figureref{fig:trans-table-all}.}
\end{table}

\parhead{The gain holds by every measure}
\begin{itemize}
\item \textbf{Across metrics.} LlamaCanon improves all $4\times6=24$ entries of \tableref{tab:trans-main}: beyond \texttt{lm-eval} and MMLU, BabiLong rises by 4.0--9.6 points, JRT accuracy rises, and perplexities fall.
\item \textbf{Run by run.} \tableref{tab:trans-main} reports the better of two learning rates, but the gain does not depend on this choice: LlamaCanon wins all 8 same-rate pairs on \texttt{lm-eval} (4 settings $\times$ 2 rates), and even the worst Canon run beats the best vanilla run by 0.6--1.3 points (\figureref{fig:trans-table-all}).
\item \textbf{Task by task.} At 8B, Canon improves 10 of the 13 \texttt{lm-eval} tasks (all but PIQA, SIQA, and COPA). On BabiLong, it raises the length-averaged score in 19 of 20 combinations of five tasks (qa1--qa5) and four settings (\figureref{fig:trans-table}).
\item \textbf{Against $2\times$ the tokens.} At 1.3B, LlamaCanon trained on 1T tokens (58.9 on \texttt{lm-eval}, 47.0 on MMLU) beats vanilla Llama trained on 2T (57.7 and 45.5).%
\footnote{The two token budgets use different batch sizes and learning-rate grids ($\{0.002,0.003\}$ at 1T, $\{0.003,0.005\}$ at 2T). We treat this cross-budget comparison as illustrative; \resultref{res:2} instead gives a target-specific, curve-based token-saving estimate.}
\end{itemize}

\begin{figure}[t!]
\centering
\includegraphics[page=4,trim={3.5bp 51.0bp 25.0bp 0.5bp},clip,width=\nipswidth{0.85}{0.88}\textwidth]{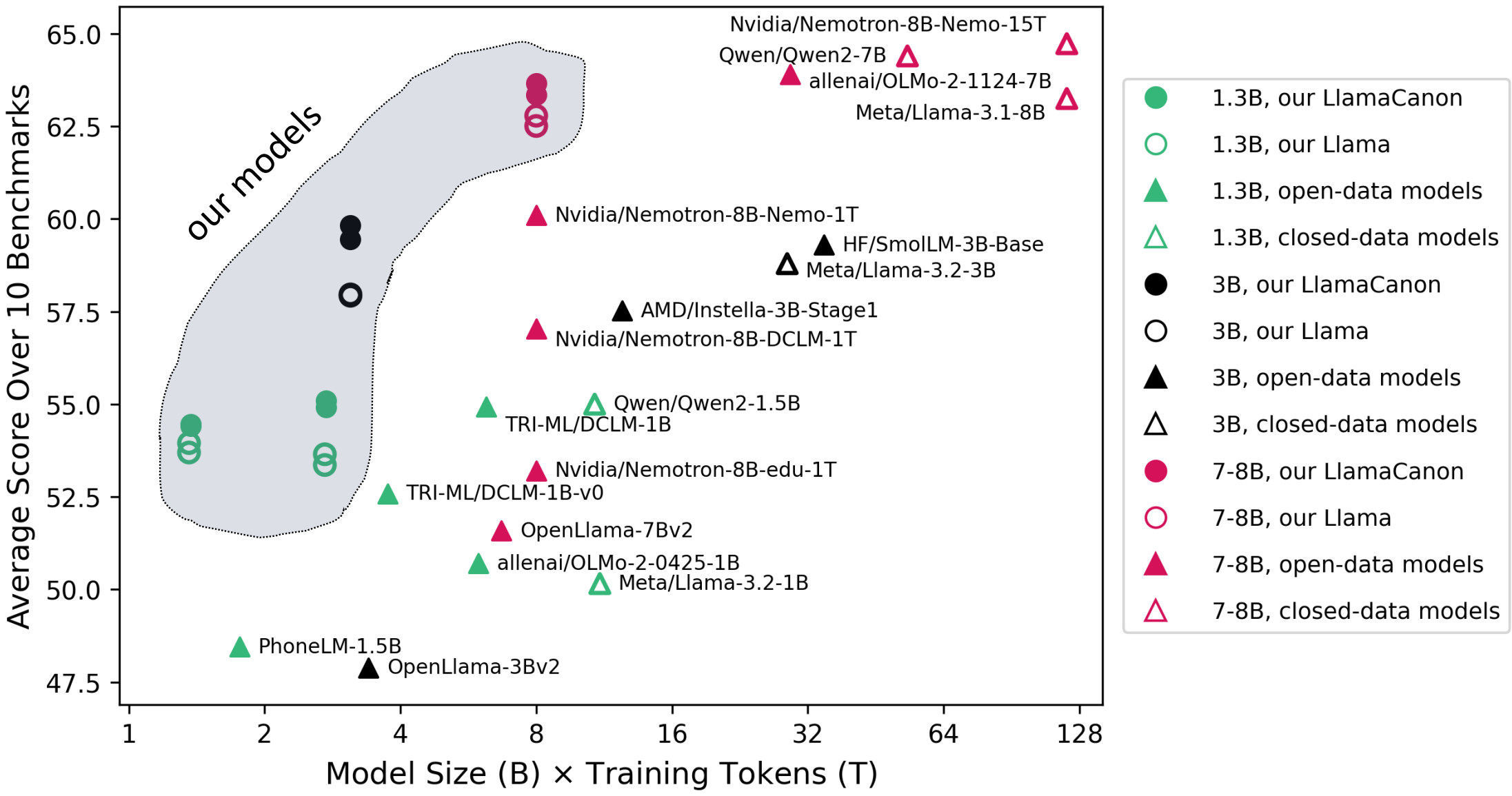}
\caption{\label{fig:scatter}\textbf{Our models are competitive with public releases trained on larger budgets.} The y-axis averages the 10 of our 13 \texttt{lm-eval} tasks that Nvidia also reports~\cite{su2025nemotroncc} (all but LAMBADA, BoolQ, and COPA). The x-axis is model size $\times$ training tokens on a log scale, where $1\,\text{B}\!\cdot\!\text{T} \approx 5200$ H200 GPU-hours. Each of our points is one run. Nvidia's points use their reported scores, since their checkpoints are unavailable; all other models are evaluated in our harness.}
\end{figure}

\parhead{\empty These are not toy baselines}
 \figureref{fig:scatter} places our models among public releases on the 10 \texttt{lm-eval} tasks Nvidia reports~\cite{su2025nemotroncc}. At 8B/1T, our vanilla Llama exceeds Nvidia's reported baseline on the same data, size, and token count by 2.5 points, and   LlamaCanon matches Llama-3.1-8B at roughly 1/15 of its model-size$\times$token budget.  The Canon gains are measured on top of competitive models, in a regime far below frontier scale but well beyond academic scale (1.3B/100B).

\parhead{Academic-scale (1.3B/100B) comparisons can hide a useful design}
Following the tradition of academic architecture studies, Part 4.1~\cite[Result 12]{Allen2025-canon} compared these architectures at 1.3B parameters on 100B tokens of FineWeb-Edu or SlimPajama. Even though every model was carefully tuned, most architectural differences vanished in that setting, and the gains from Canon, for Transformers and linear models alike, were generally smaller than the variation across random seeds. In this paper, with at least $10\times$ as many tokens and a better corpus, every matched Transformer comparison favors Canon, as the synthetic pretraining playground had predicted, although we cannot say which of the two factors, more tokens or better data, made the difference. The earlier null result was therefore not evidence against Canon: the architecture is the same in both studies, and only the experiment became sensitive enough to detect its gain.

\subsection{Result 2: Canon Saves About 30\% of Pretraining Tokens at 8B}
\label{sec:res2}

\begin{figure}[t!]
\centering
\vspace{-3mm}\includegraphics[width=0.94\textwidth]{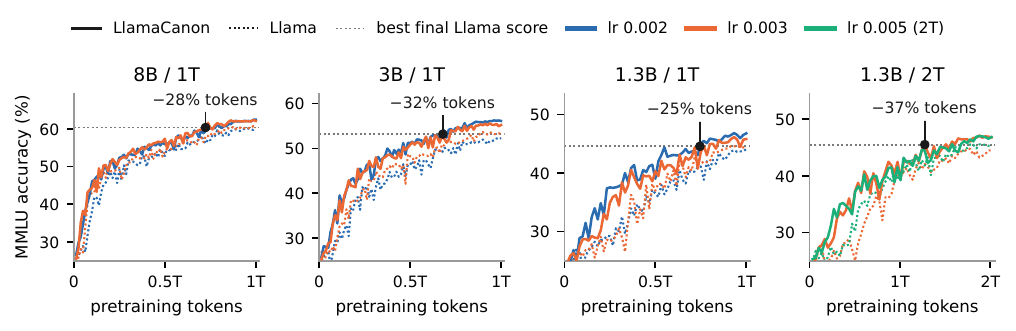}
\caption{\label{fig:curve-mmlu}\textbf{MMLU accuracy during pretraining, all 16 Transformer runs.} Solid curves add Canon, dotted curves are vanilla Llama, and colors mark the learning rate. The dot marks where LlamaCanon, taking the better of its two learning rates at each checkpoint, first reaches Llama's best final accuracy (grey dotted line), and the label gives the fraction of tokens saved.}
\end{figure}

\parhead{Compare learning trajectories, not just endpoints}
Like \tableref{tab:trans-main}, standard result tables report a single final number per model. Such a number can mislead, because benchmark accuracy is bumpy over training: MMLU fluctuates by 2--3 points between checkpoints, and on ARC-challenge the two curves interleave, so it takes a 5-checkpoint moving average to tell them apart. An endpoint difference can thus reflect \emph{only} where training \emph{happened} to stop, yet architecture comparisons \emph{commonly rest} on such final-checkpoint numbers. We therefore read the full trajectories (\figureref{fig:curve-mmlu}), which show whether one model stays ahead throughout training. As a by-product, the trajectories also give a summary that compares more easily across sizes and studies than ``$+1.8$ MMLU points'': the fraction of tokens saved to reach the same accuracy.

\begin{mdframed}
\begin{sresult}{2}[\figureref{fig:curve-mmlu}]
\label{res:2}
At 8B/1T, \defem{LlamaCanon first reaches vanilla Llama's best final MMLU accuracy with about 30\% fewer  pretraining  tokens}. It is 25--32\% in all three 1T settings and shows no sign of shrinking with scale.
\end{sresult}
\end{mdframed}

\parhead{From accuracy gains to tokens saved}
We measure the saving on MMLU, a widely reported test of knowledge and problem solving, rather than on pretraining loss, which averages over the whole corpus instead of measuring any particular skill. All loss curves are still released as interactive charts (\appendixref{app:repro}). In \figureref{fig:curve-mmlu}, which plots every run, Canon reaches vanilla's final accuracy roughly 30\% earlier in every setting. To make this precise, we keep the better of each model's two learning rates at every checkpoint and find where Canon first reaches vanilla's final accuracy (the dots). The tokens saved are
$$
\textbf{8B: } \sim\!28\%, \qquad
\textbf{3B: } \sim\!32\%, \qquad
\textbf{1.3B: } \sim\!25\%~(\text{1T}) \text{ and } \sim\!37\%~(\text{2T}),
$$
which we round to $\sim$30\% at 8B. \appendixref{app:token-details} gives the detailed computation, including its sensitivity to smoothing and to the target accuracy.

\parheadno{How large is a 30\% saving?}
Many methods report savings of 30\% or more, but such gains often shrink once the baseline is well tuned. For instance, new optimizers have been claimed to save 29--50\% of training compared with AdamW, yet against well-tuned baselines, \citet{wen2025fantastic} found that this reduces to at most 9\% for matrix optimizers at 1.2B parameters on 200B tokens. Our $\sim$30\% saving holds at 8B/1T, a model-size$\times$token budget about $33\times$ larger, against a baseline with its own learning-rate sweep. This is not a head-to-head comparison: that study measures validation loss and tests optimizers, whereas we measure MMLU and test an architecture. Three model sizes also cannot establish a scaling law, but across them our saving shows no sign of shrinking, even at the largest scale we could afford.

\section[Linear Models: Local Mixing Rivals More Elaborate Recurrent Updates]{Linear Models: Local Mixing Rivals\arxivbreak More Elaborate Recurrent Updates}
\label{sec:linear-results}

\subsection[Result 3: GLA5 and GDN2 --- Low-Rank Output Gates and Strict Alignment]{Result 3: GLA5 and GDN2 ---\arxivbreak Low-Rank Output Gates and Strict Alignment}
\label{sec:linear-upgrade}

Historically, GLA came first, and later linear models often report gains over it as the simplest baseline. Yet GLA's defaults were set early and rarely revisited, and they differ from GDN's in parameter and memory allocation, not only in recurrent memory design. Judging GDN's delta rule therefore requires both GDN and its GLA baseline at their strongest, under the same resources.

\begin{mdframed}
\begin{sresult}{3}[Eq.~\ref{eqn:gate-new}, \tableref{tab:config}]
\label{res:3}
We introduce \defem{GLA5 and GDN2}, which replace the full-rank output gate of gated linear attention and Gated DeltaNet (up to 1/3 of the recurrent layer's parameters) with a low-rank one and give the saved parameters to the rest of the layer (\sectionref{sec:linear-upgrade:gate}). We \defem{strictly align} GLA5, GDN2, and Mamba2 in parameter split, recurrent memory, and heads (\sectionref{sec:linear-upgrade:split}), so comparing them largely isolates their recurrent memory designs.
\end{sresult}
\end{mdframed}

\subsubsection{Step 1: Equalize the Parameter Split and the Recurrent Memory}
\label{sec:linear-upgrade:split}

By default, GLA~\cite{yang2023gated} spends $4d^2$ parameters per block on the recurrent layer and $8d^2$ on the MLP, whereas GDN~\cite{yang2024gated} spends $6d^2$ on each, where $d$ is the hidden width. Mamba2~\cite{dao2024transformers} has no MLP and uses $2\times$ the depth instead. Their default recurrent states also differ in size.%
\footnote{With the default configurations of the FLA library~\cite{yang2024fla} at $d=2048$, the main recurrent state per block holds $256d$ entries in GLA, $384d$ in GDN, and $512d$ in Mamba2 ($256d$ in each of its two layers per block); \appendixref{sec:setup:align} derives these numbers.}

We therefore give GLA, GDN, and Mamba2 the same nominal budgets: $6d^2{+}6d^2$ for the recurrent layer and MLP per block, a recurrent state of $256d$ entries per block, and common head counts and dimensions. Each block thus matches our Transformer blocks in size ($12d^2$, or $13d^2$ at 8B following Llama-3.1-8B, where the linear split becomes $6d^2{+}7d^2$), so all models share the same depth (\tableref{tab:config}). We call this \emph{strict alignment}: beyond sharing data, data order, and model shape, strictly aligned models have the same parameter split, recurrent memory, and heads in every layer, and train with the same context length, batch size, and number of steps. For Mamba2, strict alignment adds an MLP and halves the depth, so our Mamba2 conclusions concern this strictly aligned variant (\appendixref{sec:setup:align}). The low-rank output gate in Step~2 applies only to GLA and GDN.\looseness=-1

\subsubsection{Step 2: Reallocate the Full-Rank Gate's Parameter Budget}
\label{sec:linear-upgrade:gate}

An output gate rescales the recurrent layer's output channel by channel. Let $d_v$ be the total value dimension across all heads. For layer input $h\in\mathbb{R}^d$ and output $o\in\mathbb{R}^{d_v}$, the original GLA/GDN gate is
\begin{equation}\label{eqn:gate-old}
o \;\longmapsto\; \mathrm{RMSNorm}_{\mathrm{head}}(o) \odot \mathrm{silu}(G h), \qquad G \in \mathbb{R}^{d_v\times d}.
\end{equation}
The normalization acts within each value head, and the full-rank $G$ costs $dd_v$ parameters.

\parheadno{Why simplify this gate?}
At first sight, this output gate resembles the popular gated MLP SwiGLU~\cite{shazeer2020glu}: one branch passes through SiLU and multiplies the other elementwise. In SwiGLU, however, both branches are linear projections of the same hidden state, which the optimizer may find easier to align. Here, a projection of the input $h$ instead multiplies the output $o$ of a complex recurrent computation over the whole prefix. Coordinating the two is much harder, so we expect the learned gate to do only simple things, such as rescaling the output or suppressing a few directions, for which a low-rank gate should suffice. This reflects a central question of deep learning:
\begin{quote}
\textbf{\emph{Existence is cheap; learnability is everything.}} An architecture may express a computation that training never finds; what matters is what the optimizer learns.%
\footnote{This distinction between expressive capacity and learned behavior also guides the synthetic playground~\cite{Allen2025-canon}; in theory too, representing a target in a hierarchical network does not by itself guarantee efficient learning~\cite{allen2020backward}.}
\end{quote}

We therefore replace $G$ with a low-rank factorization, as GLA already does for its decay gate~\cite{yang2023gated}, while keeping the normalization and SiLU activation:
\begin{equation}\label{eqn:gate-new}
o \;\longmapsto\; \mathrm{RMSNorm}_{\mathrm{head}}(o) \odot \mathrm{silu}(G_1 G_2 h + b),
\qquad G_2 \in \mathbb{R}^{r \times d},\; G_1 \in \mathbb{R}^{d_v \times r},\; b \in \mathbb{R}^{d_v},\; r \ll d .
\end{equation}
The projection now costs $r(d+d_v)+d_v$ parameters, including the bias, and our 1.3--8B experiments use $r=16$. To keep the projection's initial output scale, we draw both factors with standard deviation $\sigma^{1/2}r^{-1/4}$, where $\sigma$ is that of the original $G$, so that $G_1G_2$ matches the entrywise variance of $G$, and we initialize $b$ to zero (\appendixref{app:gate-details}).

\parhead{A small change frees 1/3 of the recurrent layer}
With $d_v=2d$ in our strictly aligned models, a full-rank gate takes $2d^2$ parameters per block, 1/3 of the recurrent layer's $6d^2$ budget, whereas the rank-16 gate takes under $1\%$ as many (6.55M versus 1.07B parameters at 8B). Rather than making the model smaller, we give the saved parameters to the rest of the recurrent layer.

The resulting \bblue{GLA5} and \bblue{GDN2} proved stronger than their originals, so we use them in all large-scale GLA/GDN comparisons.%
\footnote{We found them stronger first in the synthetic pretraining playground ($\leq$100M parameters), and then in real-life ablations at 1--3B that compared them with the original GLA and GDN over the early part of training; neither set of runs is reported here. The large-scale runs evaluate low-rank gating and parameter reallocation together, and we do not measure the learned gate rank (\appendixref{app:gate-details}). GLA5 and GDN2 are internal codenames; GLA2--GLA4 are reserved for other purposes.}

\parhead{These are strong linear-model baselines}
\tableref{tab:linear-main} summarizes all 24 configurations. On the standard \texttt{lm-eval} average, almost every linear model matches or beats vanilla Llama. This agrees with the original GLA, Mamba2, and GDN studies, which report linear models competitive with or better than Llama-style Transformers~\cite{yang2023gated,dao2024transformers,yang2024gated}, but now in a carefully tuned and much larger setting. With Canon, GLA5 and GDN2 beat even LlamaCanon on \texttt{lm-eval} at every size. We return to this table several times in the following results.

\begin{table}[t!]
\centering
\small\setlength{\tabcolsep}{5pt}
\renewcommand{\arraystretch}{0.96}\begin{tabular}{@{}llcccccc@{}}
\toprule
& & \multicolumn{4}{c}{\textbf{Accuracy} $\uparrow$} & \multicolumn{2}{c}{\textbf{Perplexity} $\downarrow$} \\
\cmidrule(lr){3-6}\cmidrule(l){7-8}
& \textbf{model} & \texttt{lm-eval}(13) & \textbf{MMLU} & \textbf{JRT}(12) & \textbf{BabiLong}(20) & \textbf{Wiki ppl} & \textbf{LMB ppl} \\
\midrule
\multirow{8}{*}{\rotatebox{90}{\textbf{8B / 1T}}}
 & Llama            & 66.9 & 60.6 & \underline{64.7} & 51.3 & 9.60 & 4.60 \\
 & \ \ $+$Canon     & 67.7 & 62.4 & \textbf{64.8} & \textbf{55.3} & \textbf{9.19} & \textbf{4.20} \\
\addlinespace[2pt]
 & Mamba2           & 67.3 & 61.6 & 61.4 & 39.2 & 9.36 & 4.30 \\
 & \ \ $+$Canon     & 67.2 & 61.6 & 60.4 & 42.3 & 9.32 & 4.41 \\
\addlinespace[2pt]
 & GLA5             & 67.6 & 61.3 & 61.0 & 44.8 & 9.47 & 4.63 \\
 & \ \ $+$Canon     & 68.0 & 63.0 & 61.7 & 42.2 & 9.22 & 4.23 \\
\addlinespace[2pt]
 & GDN2             & 68.0 & 63.1 & 61.3 & 42.7 & 9.38 & 4.23 \\
 & \ \ $+$Canon     & \textbf{68.2} & \textbf{64.0} & 62.1 & 44.8 & 9.20 & 4.25 \\
\midrule
\multirow{8}{*}{\rotatebox{90}{\textbf{3B / 1T}}}
 & Llama            & 62.3 & 53.2 & \underline{58.1} & 42.4 & 11.45 & 5.47 \\
 & \ \ $+$Canon     & 63.8 & 56.1 & \textbf{59.8} & \textbf{49.4} & \textbf{11.09} & 5.09 \\
\addlinespace[2pt]
 & Mamba2           & 62.4 & 53.5 & 54.7 & 37.9 & 11.36 & 5.23 \\
 & \ \ $+$Canon     & 63.6 & 55.7 & 55.9 & 38.0 & 11.22 & 5.04 \\
\addlinespace[2pt]
 & GLA5             & 63.0 & 53.7 & 55.4 & 41.3 & 11.47 & 5.50 \\
 & \ \ $+$Canon     & \textbf{64.2} & 56.9 & 56.6 & 42.6 & 11.14 & 5.12 \\
\addlinespace[2pt]
 & GDN2             & 63.7 & 56.5 & 56.0 & 40.7 & 11.35 & 5.08 \\
 & \ \ $+$Canon     & \textbf{64.2} & \textbf{58.2} & 56.7 & 42.4 & 11.16 & \textbf{4.84} \\
\midrule
\multirow{8}{*}{\rotatebox{90}{\textbf{1.3B / 1T}}}
 & Llama            & 57.7 & 44.5 & \underline{56.6} & 36.0 & 13.67 & 7.54 \\
 & \ \ $+$Canon     & 58.9 & 47.0 & \textbf{57.8} & \textbf{45.3} & \textbf{13.17} & 7.07 \\
\addlinespace[2pt]
 & Mamba2           & 58.1 & 45.4 & 52.9 & 37.0 & 13.65 & 7.43 \\
 & \ \ $+$Canon     & 58.1 & 43.1 & 52.7 & 37.9 & 13.56 & 7.33 \\
\addlinespace[2pt]
 & GLA5             & 56.9 & 44.8 & 52.5 & 38.4 & 14.04 & 8.23 \\
 & \ \ $+$Canon     & \textbf{59.7} & \textbf{50.6} & 54.5 & 40.5 & 13.38 & 7.16 \\
\addlinespace[2pt]
 & GDN2             & 58.0 & 47.2 & 53.8 & 37.6 & 13.76 & 7.25 \\
 & \ \ $+$Canon     & 59.4 & 50.2 & 54.3 & 38.7 & 13.47 & \textbf{6.72} \\
\bottomrule
\end{tabular}
\caption{\label{tab:linear-main}\textbf{With Canon, GLA5 and GDN2 become close on \texttt{lm-eval}, while Transformers lead on JRT.} Each task takes the best of 2--3 learning rates before averaging, so a row can combine runs. Bold marks the best at each size. Underlined is vanilla Llama's JRT score, which already exceeds every linear model's, with or without Canon (\sectionref{sec:res6:suites}). Per-task scores are in \figureref{fig:linear-table}, and every individual run in \figureref{fig:linear-table-all}.}
\end{table}

\subsection{Result 4: With Canon, GLA5 Matches GDN2}
\label{sec:res4}

Each head in GLA and GDN keeps a matrix memory $S_t$, which the query reads out as $o_t=S_tq_t$. The two models differ mainly in how they update $S_t$ from the current token's key $k_t$ and value $v_t$. GLA~\cite{yang2023gated} uses gated accumulation, arguably the simplest form of linear attention: it decays the old memory and adds the new key--value pair. GDN~\cite{yang2024gated} adds a delta rule: it first reads what the memory returns for $k_t$ and then writes only the correction toward $v_t$,
$$
\text{GLA:}\quad S_t = S_{t-1}\,\mathrm{Diag}(\alpha_t) + v_t k_t^{\top},
\qquad\qquad
\text{GDN:}\quad S_t = \alpha_t S_{t-1} + \beta_t\,\big(v_t - \alpha_t S_{t-1}k_t\big)\,k_t^{\top},
$$
where the decay $\alpha_t$ (a vector in GLA, a scalar in GDN) and the write strength $\beta_t\in(0,1)$ are computed from the current token. Such additions to gated accumulation, the delta rule in GDN and the selective state-space update in Mamba2, are widely used: GDN and Mamba2 underpin some of the largest open hybrid models,  from early ones such as Qwen3-Next~\cite{Qwen3next} and Falcon-H1~\cite{tiifalconh1}, respectively, to many that followed. 

Without Canon, GDN2 scores above GLA5 on the standard benchmark average at every size (\tableref{tab:linear-main}). Because the two models are strictly aligned (\resultref{res:3}), this lead  \arxivem{truly reflects}  GDN's recurrent memory design. \arxivbf{But} does it survive once both have Canon at A, C, and D?%
\footnote{Every linear model here already has a short convolution at position B (Canon-b; \sectionref{sec:canon-recap}). The original GLA paper did not use one, and the FLA implementation~\cite{yang2024fla} added it later; we compare against this stronger version, since GLA without it performs poorly~\cite{Allen2025-canon}.}

\begin{figure}[t!]
\centering
\begin{minipage}[c]{\nipswidth{0.46}{0.56}\textwidth}
\centering
\includegraphics[width=\textwidth]{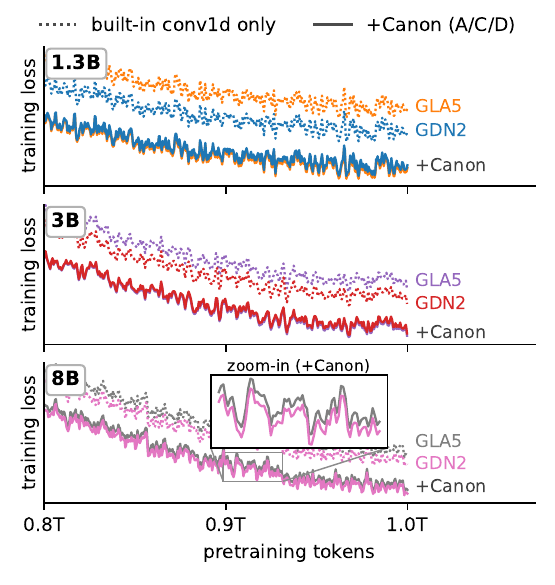}
\end{minipage}%
\hfill
\begin{minipage}[c]{\nipswidth{0.51}{0.41}\textwidth}
\captionsetup{format=plain}
\caption{\label{fig:loss-linear}\textbf{With Canon, GLA5 and GDN2 have nearly identical training curves.} Training loss over the last 20\% of pretraining at learning rate 0.003, for 1.3B, 3B, and 8B from top to bottom. Dotted curves use only the built-in convolution, and solid curves add Canon at A, C, and D. Panels have separate y-scales. The zoom-in enlarges a short stretch of the two 8B runs with Canon, showing that even their small rises and falls line up. As in the released interactive charts, each curve is a trailing moving average over 50 logged values.}
\end{minipage}
\end{figure}

\begin{mdframed}
\begin{sresult}{4}[\tableref{tab:linear-main}, \figureref{fig:loss-linear}, \figureref{fig:curve-linear}]
\label{res:4}
With Canon, GDN2's lead disappears: \defem{GLA5 and GDN2 become almost identical}, both in training loss  throughout pretraining and  on the standard benchmark average (within 0.3 points). GLA5 with Canon also reaches the final MMLU accuracy of GLA5 without it using about 24\% fewer tokens at 8B, and saves more at smaller sizes.
\end{sresult}
\end{mdframed}

\parhead{Canon closes the standard-benchmark gap}
Without Canon, GDN2 leads GLA5 on \texttt{lm-eval} at every size, by up to 1.1 points. With Canon, the two are within 0.3 points of each other at every size, although GDN2 keeps a 1-point MMLU lead at 8B (\tableref{tab:linear-main}). While this near-tie may surprise first-time readers, the synthetic pretraining playground~\cite[Results 9--10]{Allen2025-canon} had already predicted it with networks under 100M parameters.

\parhead{\empty Surprisingly, their training curves coincide}
The agreement holds throughout training, not just at the end. Without Canon, GLA5's training loss stays above GDN2's at every size, whereas with Canon the two curves nearly coincide. Because the two models see identical data batches in identical order, even the small rises and falls of their curves line up (\figureref{fig:loss-linear}). Even their benchmark accuracy curves over training are very much alike: \appendixref{app:linear-curves} compares MMLU as an example, and the full trajectories of every benchmark are open-sourced (\appendixref{app:repro}). This challenges what GDN's recurrent design actually contributes: does its delta rule truly correct what is already stored, or does it merely make the three preceding tokens easier to retrieve --- exactly what Canon provides?

\parhead{Canon also saves tokens for linear attention}
We measure the saving as in \resultref{res:2}. GLA5 with Canon reaches its baseline's final MMLU accuracy using \arxivbblue{24\%} fewer tokens at 8B, \arxivbblue{40\%} fewer at 3B, and \arxivbblue{62\%} fewer at 1.3B.  The saving shrinks with size but stays substantial at 8B, even though GLA5 already has a built-in convolution.

\subsection{Result 5:  Horizontal Mixing Dominates Architectural Sophistication}
\label{sec:res5}

\begin{figure}[t!]
\centering
\includegraphics[page=3,trim={0bp 245.8bp 422.4bp 0bp},clip,width=0.85\textwidth]{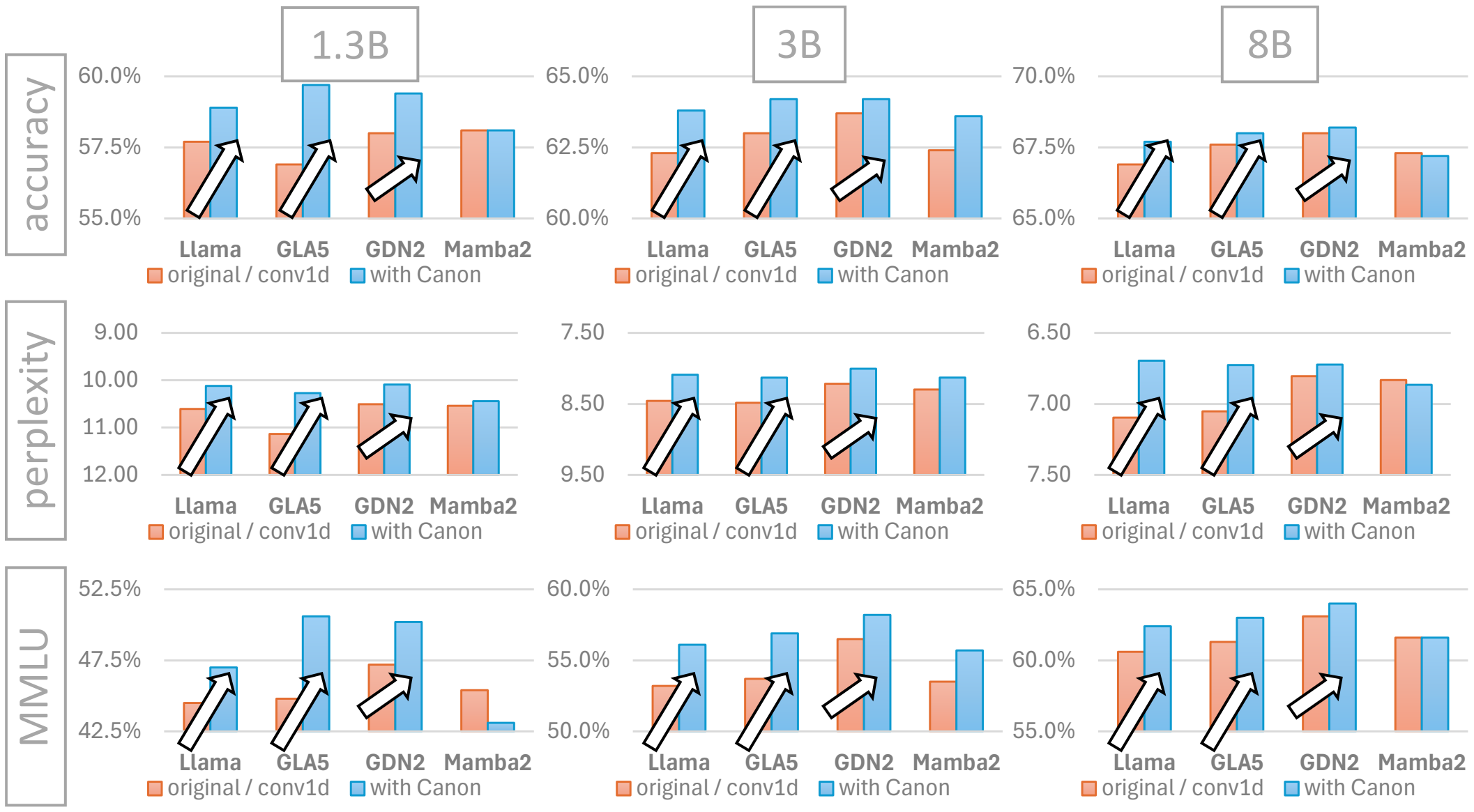}
\caption{\label{fig:marginal}\textbf{Canon helps most where an architecture's own local mixing is weakest.} Each pair of bars compares a model without Canon (orange; the linear models keep their built-in convolution) and with Canon (blue), each at the best of 2--3 learning rates (\tableref{tab:linear-main}). Rows show the \texttt{lm-eval} average over 13 tasks, the mean of WikiText and LAMBADA perplexity (axis inverted, so up is better), and MMLU. Arrows mark the overall pattern: Transformers and GLA5 gain most, GDN2 about half as much, and Mamba2 changes little except at 3B.}
\end{figure}

\resultref{res:4} compared GLA5 and GDN2 head to head. Here we ask how much Canon helps each architecture over its own baseline, with the three linear families strictly aligned (\figureref{fig:marginal}). If the gains differ, the ranking before Canon is added partly reflects how much local mixing each architecture already has built in.

\begin{mdframed}
\begin{sresult}{5}[\figureref{fig:marginal}]
\label{res:5}
\defem{Canon helps most where an architecture's own local mixing is weakest}: the Transformer and GLA5 gain most, GDN2 gains about half as much as GLA5, and Mamba2 gains little. At every size,  Canon's \texttt{lm-eval} gain\ for GLA5 covers GDN2's entire lead over GLA5 when neither has Canon,  suggesting that the extra machinery of GDN2 and Mamba2 is largely horizontal mixing in disguise. \end{sresult}
\end{mdframed}

\parhead{Why the gains differ}
Canon supplies information directly from neighboring tokens, so it should help most where an architecture does little local mixing on its own. The Transformer and GLA5 gain most overall. The Transformer has no local mixing of its own, and GLA5's gated accumulation is the simplest recurrent memory design. GDN2 gains about half as much as GLA5 at every size, and Mamba2 gains least, with a clear gain only at 3B (\figureref{fig:marginal}). The delta rule and the selective SSM update thus appear to supply some of this local transfer already.\looseness=-1

\parhead{The pattern holds throughout training}
Canon's gains do not depend on where training stopped or on picking each task's best learning rate. At every size and learning rate, Canon lowers training loss most for GLA5 and least for Mamba2, whichever late-training window we average over (\appendixref{app:linear-curves}). The released benchmark curves show the same broad pattern throughout training (\appendixref{app:repro}), and the final scores for individual runs show it too (\figureref{fig:linear-table-all}). Two benchmark scores decline with Canon (\tableref{tab:linear-main}). At 1.3B, Mamba2 with Canon stays 2--3 MMLU points below its baseline over the whole second half of training, a genuine regression, although its \texttt{lm-eval} average is unchanged. At 8B, adding Canon to GLA5 lowers its BabiLong score by 2.6 points, but BabiLong is noisy and the linear models already struggle on it (\sectionref{sec:res6:suites}).

\parhead{\empty What the extra machinery actually does}
Canon adds under 0.2\% to the parameters, yet at every size its \texttt{lm-eval} gain for GLA5 is at least as large as the entire lead GDN2 held over GLA5 without Canon (\tableref{tab:linear-main}). On standard benchmarks at this scale, much of what the delta rule and the selective SSM add therefore appears to be short-range horizontal mixing, and a comparison run without local mixing on both sides can credit a gain to the wrong component. These mechanisms are perhaps not worthless: they do help architectures that lack Canon-like local mixing, which remain the common case while Canon is not yet widely adopted. Beyond such mixing, however, what they add may be very incremental. This is the output-gate lesson of \resultref{res:3} one level up: an elaborate mechanism that \emph{can} express rich interactions may, in training, learn little more than a simple one --- existence is cheap; learnability is everything.

\parhead{This structure is invisible at academic scale}
At 1.3B parameters and 100B tokens, a typical scale for academic architecture research, the differences in Canon gains across architectures disappear completely~\cite[Result 12]{Allen2025-canon}.%
\footnote{At that scale, the only clearly visible difference among the linear models was between GLA with and without a short convolution. Here we raise the bar: every linear model keeps its built-in convolution (Canon-b), and we ask what Canon adds at A, C, and D on top of it, a gain that was invisible at academic scale.}
In contrast, our 1T-token runs make these differences clearly measurable, and they persist at least up to 8B. The synthetic pretraining playground had already shown the same pattern at $\leq$100M parameters, so these results were predicted in advance.

\section{Deficits That Standard Benchmarks Hide}
\label{sec:beyond}

So far, the linear models look excellent: with Canon, they match or exceed Transformers on standard benchmarks, which mostly test knowledge and commonsense. This section tests two basic in-context skills that standard benchmarks \arxivem{barely measure}: retrieving information from a supplied passage (\resultref{res:6}), and the simplest latent two-hop reasoning over it (\resultref{res:7}).

\subsection{Result 6: Linear Models Cannot Reliably Retrieve from Short Contexts}
\label{sec:res6}

We start with the simpler of the two skills: recalling information that the input states explicitly.

\begin{mdframed}
\begin{sresult}{6}[\figureref{fig:hop-length}]
\label{res:6}
At 1.3--8B/1T, linear models that match Transformers on standard benchmarks \defem{cannot reliably recall information stated earlier in their input}. Recall is near-perfect without background text, but with 500 background tokens, linear models fall to 72--98\% while Transformers stay at $\geq$99.5\%, and at 2k, inside every training context, no linear model exceeds 73.3\%. Linear models improve from 1.3B to 8B: they learn this skill, but far more slowly than Transformers.
\end{sresult}
\end{mdframed}

This deficit is invisible on \texttt{lm-eval}, our average of 13 standard benchmarks of the kind most studies report: there, with Canon, GLA5 and GDN2 even score above LlamaCanon at every size (\tableref{tab:linear-main}). For example, at 8B and learning rate 0.002, GDN2 with Canon scores 67.9 on \texttt{lm-eval} against LlamaCanon's 67.1, yet with 2k background tokens these same two checkpoints retrieve a stated birth year 68.2\% and 98.2\% of the time, respectively (\figureref{fig:linear-table-all}; task details below). We do not dispute these scores: we reproduce the linear models' competitiveness on the tasks most linear-model papers report. But knowledge and commonsense benchmarks rely largely on what a model has stored in its weights, and there linear models are in fact strong: Part 4.1 found that, with Canon, they store about 40\% more knowledge per parameter than Transformers, as their recurrent structure supports associative memory, and that Mamba2 and GDN manipulate knowledge about as well as Transformers~\cite{Allen2025-canon}. On \texttt{lm-eval}, this strength masks a weakness, since recalling information from the input is a different capability. We first isolate it with a minimal task (\sectionref{sec:res6:task}), then turn to existing retrieval tests, which only hint at the gap or miss it entirely (\sectionref{sec:res6:suites}).

\subsubsection{Five-Statement Retrieval under Interference}\label{sec:res6:task}

\begin{figure}[t!]
\centering
\includegraphics[page=1,trim={2.9bp 336.3bp 169.2bp 0bp},clip,width=\nipswidth{0.72}{0.8}\textwidth]{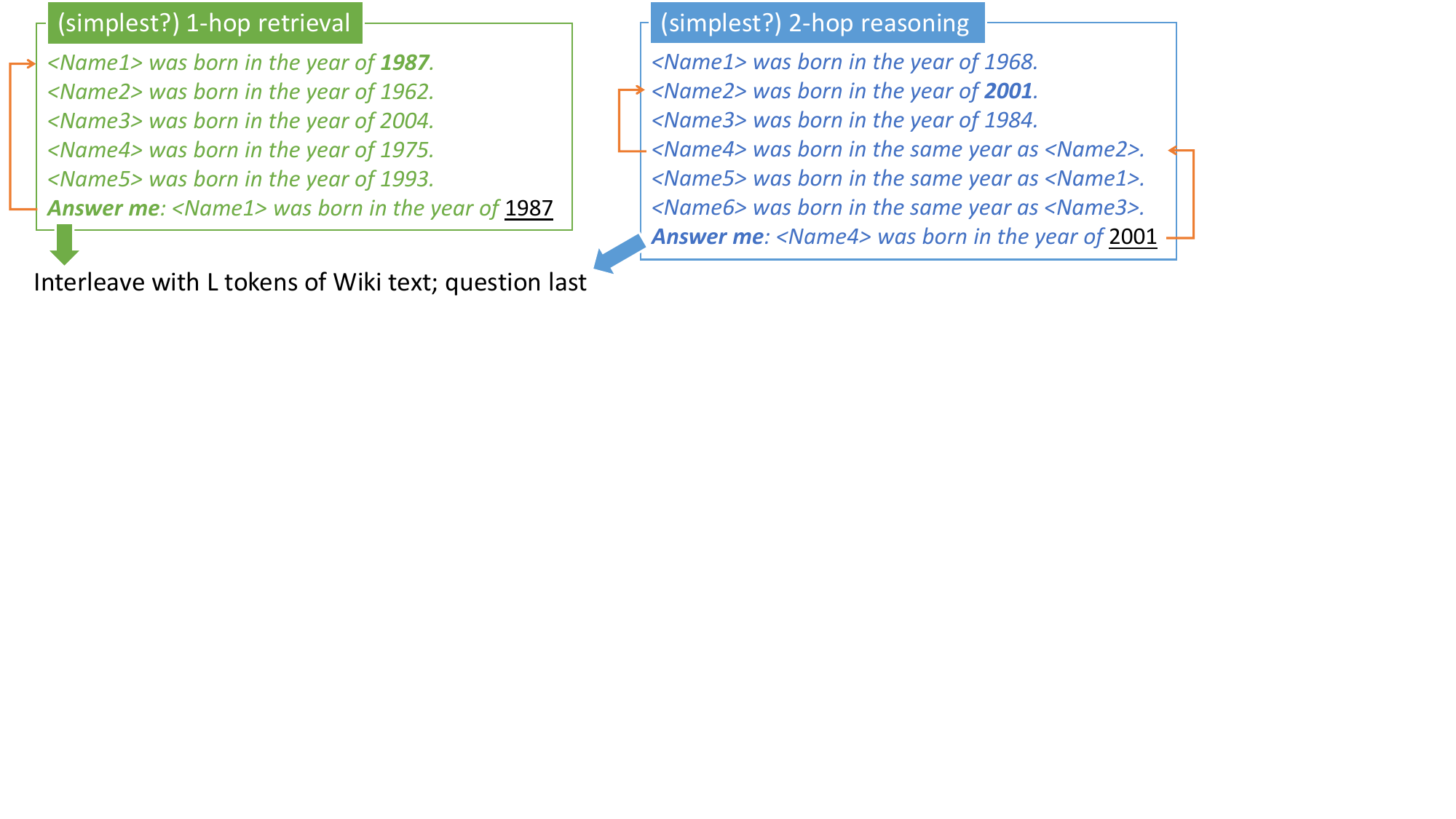}
\caption{\label{fig:hop-tasks}\textbf{\empty Our  1-hop-$L$ and 2-hop-$L$ retrieval tasks, shown at $L=0$.} In the 1-hop task (left), the model retrieves a stated birth year. In the 2-hop task (right), it must first use a same-year statement to find whose birth year to retrieve. For $L>0$, the statements are interleaved, in order, with $L$ tokens of unrelated Wikipedia text, and the question comes last. Names are random full names, and years are distinct and random in $[1950,2010)$, so guessing scores 20\% and 33.3\%. Other variants of both tasks behave similarly (\sectionref{rem:not-special}). Expected answers are underlined.}
\end{figure}

\parhead{The 1-hop-$L$ task}
The task behind \resultref{res:6} works as follows. We write five ordinary statements, each giving a person's birth year, such as ``[Name] was born in the year of [Year].'', then ask for the first person's year (\figureref{fig:hop-tasks}, left).  The answer is stated in the input, so no prior knowledge or reasoning is needed.  We interleave the statements with $L$ tokens of unrelated Wikipedia text, and $L$ is the only variable; at $L=0$, the input is just the statements and the question. We call this task 1-hop-$L$.  It is arguably the second-simplest retrieval task, after finding a single distinctive needle as in NIAH-S (\sectionref{sec:res6:suites}).  Random guessing would score 20\%. Each model answers zero-shot and generates up to 10 tokens greedily. An answer is correct if it contains the right year.%
\footnote{\appendixref{app:eval:prompts} gives the exact prompts, including the short instruction and the context wrapper of the 2-hop task.}

\parhead{Why birth years}\phantomsection\label{rem:not-special}We chose birth years on purpose. Statements like these are common in pretraining text, for example in Wikipedia biographies, so the task is as natural a retrieval test as we could make it: a model is unlikely to fail because the format is unfamiliar, or to succeed because the relevant sentence stands out, as an artificial needle does.  For this task and its two-hop extension (\sectionref{sec:res7}), we also tried other attributes, such as cities, college names, and full birth dates, as well as other question wordings and answer formats. The conclusions held: 1-hop retrieval remained imperfect for linear models, and 2-hop accuracy stayed near random guessing. Among the tests we tried, only artificial needles such as NIAH-S hide the 1-hop deficit (\sectionref{sec:res6:suites}). 

\begin{figure}[t!]
\centering
\includegraphics[width=\nipswidth{0.9}{1}\textwidth]{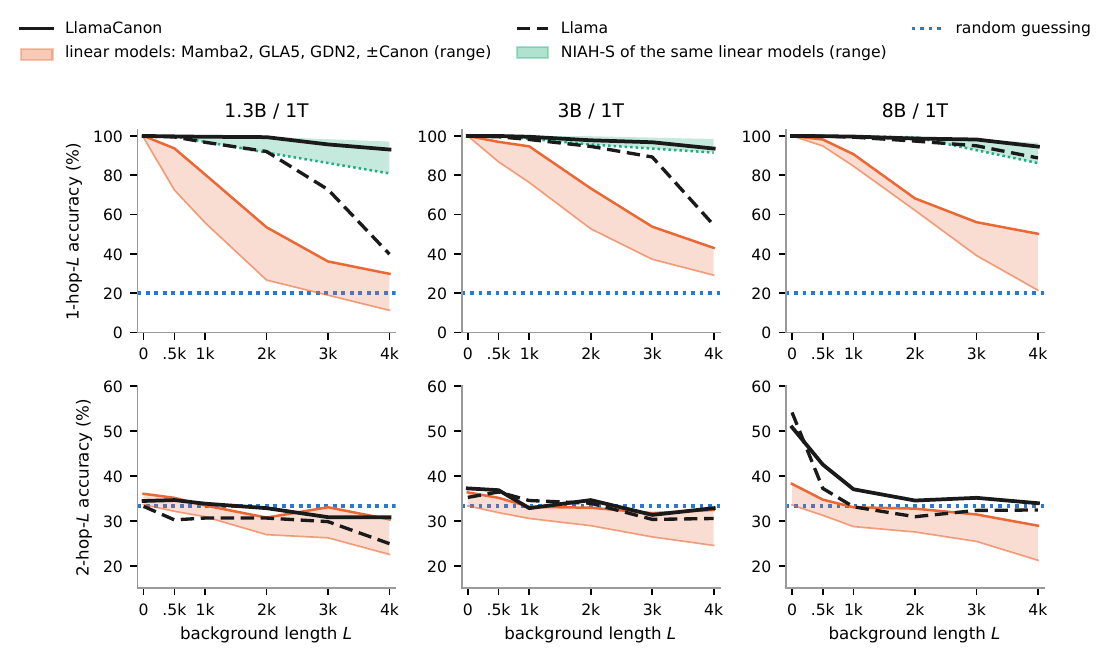}
\caption{\label{fig:hop-length}\textbf{Linear models lose retrieval reliability with only 500 background tokens, while latent two-hop reasoning stays near random guessing for almost every model.} Accuracy (\%) on 1-hop-$L$ (top) and 2-hop-$L$ (bottom) at 1T tokens, each configuration at its best of 2--3 learning rates per $L$. \textbf{Black lines} show the two Transformers. The {\color{figorange}\textbf{orange band}} spans the six linear configurations (Mamba2, GLA5, and GDN2, with and without Canon), and the {\color{figgreen}\textbf{green band}} spans these six models on NIAH-S, averaged over three variants. {\color{figblue}\textbf{Dotted lines}} mark random guessing. For each configuration and length, see \figureref{fig:linear-table}.}
\end{figure}

\needspace{4\baselineskip}
\parhead{The gap opens early and widens with background length}
\begin{itemize}
\item \textbf{$L=0$.} All 24 configurations score 99.1--100\%, so any later decline reflects interference from the background, not an inability to answer.
\item \textbf{$L=500$.} Transformers stay at 99.5--100\%, while linear models fall as low as \bblue{72.5\%} at 1.3B; even at 8B, they score only 94.9--98.2\%.
\item \textbf{$L=2$k.} Even at 8B/1T, the best linear model reaches only 68.2\%, while both Transformers stay above 97\%; at no size does a linear model exceed 73.3\%.
\item \textbf{$L=4$k.} The best 8B linear model reaches only 50.2\%, against LlamaCanon's 94.6\%, even though 4k tokens is only about 6 pages of text.
\end{itemize}
At every size and every nonzero length in \figureref{fig:hop-length}, even the weaker Transformer (i.e., without Canon) outscores the best linear model.
Context length cannot explain this gap: both 500 and 2k background tokens lie well inside every training window, and the 1.3B and 3B linear models were \textbf{\emph{even trained on 16k- and 12k-token contexts}}, against 4k for the Transformers. Our setup, if anything, favors the linear models in two more ways: at the same token budget they take two to three times as many optimizer steps, and at 1.3B and 3B they get one more learning rate to choose from (\sectionref{sec:setup}). They still show the deficit.

\parhead{The delta rule does not close the gap}
GDN's delta rule targets in-context recall, yet gives no consistent edge on 1-hop-$L$: at 2k background tokens, GDN2 with Canon reaches 47.9\%, 54.9\%, and 68.2\% at 1.3B, 3B, and 8B, against 48.3\%, 71.4\%, and 66.6\% for GLA5 with Canon (\figureref{fig:linear-table}).

\begin{remark}[Likely not a capacity limit]\label{rem:not-capacity}
Earlier work explains why linear models and other attention-free models fail at recall in three ways: a fixed-size state too small to hold the input~\cite{jelassi2024repeat}, too little input-dependent mixing~\cite{arora2024zoology}, or limited memory that forces the model to decide what to keep before it sees the question~\cite{arora2024just}. In our task, raw memory size is unlikely to be the problem: each linear block holds $256d$ entries of recurrent memory (\appendixref{sec:setup:align}), far more than five names and their birth years need. What likely remains is learning: because the question comes last, a linear model must learn to keep these pairs while background text streams through its state, whereas a Transformer keeps every token and need not learn this. We also do not rule out that much longer training (say, 10T tokens) or a better pretraining data mix would let linear models retrieve reliably within 4k tokens. Our point is that they learn this skill slowly: with the same training budget, Transformers learn it far better.
\end{remark}

\parhead{The Canon recipe improves retrieval and length generalization}\phantomsection\label{sec:res6:canon-retrieval}
LlamaCanon improves Transformer accuracy on 1-hop-$L$ with 4k background tokens. At 1.3B, for example, accuracy rises from 39.8\% for vanilla Llama to 93.1\% for LlamaCanon. Adding Canon generally helps GDN2 and GLA5 on 1-hop-$L$ as well (\appendixref{app:one-hop-results}). With 5k background tokens, beyond the Transformers' training window, vanilla Llama's accuracy falls to 0.0--2.6\%, while LlamaCanon retains 26.7--70.2\%.

\subsubsection{Existing Retrieval Tests Hint at the Gap or Miss It}
\label{sec:res6:suites}

\parhead{JRT and BabiLong hint at the retrieval gap}
\bblue{JRT}~\cite{arora2024just} asks models to answer questions about a supplied passage. For example, for a passage about the Titanic, the model may be asked who located the wreck. JRT caps prompts at 1024--2048 tokens, so they fit within every model's training window. Despite the short prompts, LlamaCanon leads the best linear model by about 3 points at every size, and even vanilla Llama scores above every linear model (\tableref{tab:linear-main}). The gap is larger on \bblue{BabiLong}~\cite{kuratov2024babilong}: at 8B, LlamaCanon reaches 55.3 against the best linear model's 44.8. Both suites thus show the deficit, though less cleanly than our 1-hop-$L$ task above: their questions also involve reading comprehension or reasoning, and BabiLong scores are noisy.%
\footnote{BabiLong is also hard: models scored near random guessing on it at academic scale (1.3B/100B), and at 8B/1T it still separates them only to a limited, task-dependent degree (\figureref{fig:linear-table}). Near-random scores, like the near-perfect ones on NIAH-S below, can obscure architectural differences.}

\parhead{Single-needle tests miss the gap}
\bblue{NIAH-S}~\cite{hsieh2024ruler} makes the linear models look nearly flawless. At 0.5k tokens, all 24 configurations, including the Transformers, score 99.4--100\% (\figureref{fig:linear-table}). At the lengths shared by both tests in \figureref{fig:hop-length}, the linear models' NIAH-S scores stay above their 1-hop-$L$ scores. Our 8B Mamba2 and GDN2, trained on 4k-token contexts, even score 100\% on NIAH-S1 at 8k, and many prior works cite such results as evidence that linear models generalize to longer contexts. Such scores are genuine but misleading, and  \emph{the reason is the needle}:  NIAH-S hides one artificial sentence, such as ``One of the special magic numbers for [word] is: 7392041.'' That sentence stands out from the surrounding text and gives attention-based and recurrent models a strong signal to keep it. Ordinary information, such as a birth year, blends into the surrounding text and offers no such cue. A near-perfect NIAH-S score therefore overstates how reliably a model recalls information from its input.

\subsubsection{Implications for Hybrids}
\label{sec:hybrid-labor}

\parhead{Hybrids work --- but perhaps not for the usual reason}
Linear models are usually promoted for long contexts, since their cost per token does not grow with input length. Hybrid models~\cite{lieber2024jamba,Qwen3next,tiifalconh1}, which mix attention and linear layers, are usually explained as a compromise: attention for quality, linear layers for efficiency. Together with Part 4.1, our results point to a different division of labor: \textbf{\emph{linear layers are good at knowledge-based and very-short-context skills, while attention is needed to recall information from the input beyond a few hundred tokens.}}%
\footnote{In Part 4.1, linear models with Canon stored about 40\% more bits per parameter than Transformers on \textsc{Capo}, a knowledge-capacity task, and Mamba2 and GDN were competitive on \textsc{Mano}, a knowledge-manipulation task; standard benchmarks largely measure such skills. The recall deficit is \resultref{res:6}.}
Because linear layers learn such recall far more slowly than attention, even much longer training is unlikely to let them replace more than sliding-window attention over a few thousand tokens, let alone attention over the very long contexts, such as 1M tokens, that they are often expected to handle. The 1-hop-$L$ retrieval gap between Transformers and linear models motivates the prediction that hybrids will need some attention until recall inside linear layers improves. \appendixref{sec:res6:hybrid} outlines a hybrid experiment that could test this prediction directly.

\subsection{Result 7: Latent Two-Hop Reasoning Barely Emerges at 8B/1T}
\label{sec:res7}

\resultref{res:6} tested 1-hop retrieval: recalling a birth year stated directly in the input. We now turn to latent two-hop reasoning in the same birth-year setting.

\parhead{The 2-hop-$L$ task}
To test latent two-hop reasoning, we extend 1-hop-$L$. We state three people's birth years, then say that each of three other people was born in the same year as one of them, and ask for one of these other people's years (\figureref{fig:hop-tasks}, right). To answer, the model must first use a same-year statement to find whose year to retrieve, and then retrieve it. This is arguably the simplest two-hop question.%
\footnote{Random guessing among the three stated years scores 33.3\%, and with only two people it would already score 50\%. A short instruction at the start of the prompt explains that some birth years are stated directly and others only through same-year statements (\appendixref{app:eval:prompts}).}
The prompt ends with ``[Name] was born in the year of'', exactly as in 1-hop-$L$, where every model scores 99.1--100\% without background text.

\begin{remark}[Latent reasoning]\label{rem:latent}
 Because the 2-hop-$L$ prompt asks for the year directly, without asking for the intermediate person or prompting a chain of thought, the task tests \emph{latent} two-hop reasoning. Asking first ``Who was born in the same year as [Name]?'' would split it into two explicit one-hop questions.  Latent reasoning and chain of thought are orthogonal: with a chain of thought of a given length, a model that takes two hops per written step could, in principle, reason twice as deep as one that takes a single hop.
\end{remark}

\begin{mdframed}
\begin{sresult}{7}[\figureref{fig:hop-length}]
\label{res:7}
Even at 8B/1T, \defem{no model reliably beats random guessing (33.3\%) on  arguably the simplest  latent two-hop question}, stated in seven short  sentences: three birth years, three same-year statements, and  the  question. Only the 8B Transformers reach about 50\% without background text. That advantage mostly vanishes once 500 background tokens are added.
\end{sresult}
\end{mdframed}

\parhead{Most models remain near random guessing}
With no background text at all, every configuration except the 8B Transformers stays within 5 points of the 33.3\% guessing reference, even at its best learning rate (\figureref{fig:hop-length}, bottom). We stress again that the task itself is \arxivem{only seven sentences, about 100 tokens}. The failure thus occurs before any distracting text is added: since 1-hop recall is near-perfect at this length, the difficulty lies in the extra hop rather than in long inputs. Raw memory size is unlikely to explain the failure either. These seven sentences fit easily in every linear model's $256d$  recurrent  state (Remark~\ref{rem:not-capacity}). The Transformers keep every token in their KV cache, yet below 8B they also stay near random guessing.

\parhead{Neither hybrids nor other task variants rescue it}
The first question this result invites is whether a hybrid would do better. Outside the 64-run suite, we explored hybrids with interleaved layers, parallel attention and recurrent branches, and learned gates across several attention fractions, but none made two-hop reasoning reliable (\appendixref{sec:res6:hybrid}). Other prompt wordings, answer formats, and two-hop variants beyond birth years did not change the picture either (\sectionref{rem:not-special}).

\begin{remark}[Latent two-hop reasoning emerges late]\label{rem:8b-fragile}
Result 7 shows that general-purpose pretraining begins to develop latent two-hop reasoning only in our largest runs (8B parameters, 1T tokens of one of the best public corpora), and even there the advantage over random guessing mostly vanishes with 500 background tokens. Larger models trained on more data may well do better. Still, this poses a real problem for architecture research: a skill this basic appears only at a scale where each run costs roughly \$100k. Even a study that can afford such a run for every design being compared would meet the same problem again with latent three-hop reasoning, which would presumably need even larger runs.
\end{remark}

\section{What the Deficits Mean}
\label{sec:lessons}

Results 6 and 7 test base models, that is, models after pretraining alone. Two practical questions follow. First, if pretraining leaves retrieval and reasoning skills weak, can post-training simply add them later (\sectionref{sec:res6:sft})? Second, if even our 8B/1T runs leave models near random guessing on latent two-hop reasoning, how can architectures be compared on such skills (\sectionref{sec:res7:lessons})?

\subsection{Post-Training Needs a Capable Starting Point}
\label{sec:res6:sft}

Practitioners have observed that \emph{once a model has finished pretraining, it can be hard to teach it new skills, such as reasoning skills, during post-training}.  Two of our earlier works explain this with controlled experiments.  In Part 2.2~\cite{YXLZ2024-gsm2}, a model pretrained only on error-free math solutions has not learned error correction, and teaching it this skill during post-training comes too late. Parameter-efficient fine-tuning such as LoRA fails, because a genuinely different skill requires larger parameter changes than such methods can make. Full-model fine-tuning also fails, unless it trains about as long as pretraining itself, which is \textbf{effectively continued pretraining} (or mid-training) rather than post-training. In Part 3.2~\cite{AL2023-knowledgeUB}, models that flexibly retrieve knowledge stored in their weights still struggle with even the simplest operations on that knowledge (e.g., classification and comparison, let alone more complex reasoning) unless they first write it down with chain of thought, turning reasoning over stored knowledge (harder) into in-context reasoning (easier). As in Part 2.2, teaching these operations after pretraining would again require continued pretraining rather than a short post-training stage. Therefore,
\begin{center}
\emph{the base model should already have the basic skills that post-training builds on.}
\end{center}
These include at least basic two-hop reasoning such as  our  2-hop-$L$ task, and ideally more complex reasoning. This agrees with the common observation that post-training works best on a strong base model: for example, reinforcement learning with verifiable rewards (RLVR) mostly sharpens reasoning paths that the base model already has~\cite{yue2025limitsrlvr}, although prolonged RL can expand them~\cite{liu2025prorl}.

\parhead{Prepare pretraining data for in-context reasoning earlier}
Many training pipelines reserve most explicit reasoning data for mid- or post-training. In such a pipeline, the pretrained model may still lack basic reasoning skills: in our 8B models, even the simplest latent two-hop reasoning only begins to emerge after 1T pretraining tokens (Remark~\ref{rem:8b-fragile}). This may be too late: later stages would then have to reshape the model for reasoning after it has absorbed most of its knowledge. We believe it is better for the model's internal states to start preparing for reasoning while that knowledge is being acquired. For practitioners, we therefore recommend including in-context reasoning examples (\emph{real or synthetically generated}), such as retrieval and multi-hop questions, in pretraining data from an early stage and interleaved with the rest of the data, rather than reserving them for mid-training, let alone post-training. Even simple two-hop examples like ours should appear well before the 1T-token mark; harder material such as competition-level mathematics can still come later and build on this foundation. In this sense, \bblue{reasoning data should connect pre-, mid-, and post-training}.

\subsection{Architecture Comparisons Need Measurable Skills}
\label{sec:res7:lessons}

To compare architectures on a skill, an experiment must first be able to measure that skill. Our results show two ways in which real-life pretraining can fall short. First, standard benchmark averages such as \texttt{lm-eval} can hide retrieval gaps between same-sized models that appear with only 500 background tokens (\resultref{res:6}). Second, nearly every model stays near random guessing on arguably the simplest latent two-hop question (\resultref{res:7}). Such similarly low scores cannot tell us which architecture is better at learning this skill. Worse, since these models have barely acquired even latent two-hop reasoning, \emph{any} real-life benchmark comparison among them likely involves almost no two-hop reasoning, whatever the benchmark is meant to test.

Part 4.1~\cite[Result 12]{Allen2025-canon} warned that in academic-scale comparisons, architectural differences can all but disappear behind seed variation and near-random reasoning scores. Scaling up makes the Canon gain visible (\namedref{Results}{res:1}, \ref{res:2}, \ref{res:4}, and~\ref{res:5}), but near-random reasoning scores persist even at 8B (Remark~\ref{rem:8b-fragile}), so real-life pretraining still misses many architectural differences. Scaling up is no practical way out: latent two-hop reasoning barely emerges at 8B/1T, so smaller studies of architectures or optimizers cannot measure it at all, and deeper latent reasoning would need runs far too costly to train once per architecture or optimizer. When two architectures score within a point of each other, the honest reading is often not ``they are equivalent'' but ``this experiment cannot tell them apart''.

The synthetic pretraining playground of Part 4.1 avoids these measurement problems by design. For each skill, such as latent multi-hop reasoning, its pretraining data is designed to be rich enough for models to learn that skill well, and the skill is then tested on new inputs. Architectures can thus be compared skill by skill, and their differences in \emph{learning} each skill become measurable below 100M parameters. For example, on \textsc{Depo2}, a synthetic multi-hop retrieval task with multi-token names, a Transformer with Canon quickly reaches near-perfect accuracy, while Mamba2, GLA, and GDN with Canon struggle to reach 99\% accuracy even at one or two hops on the same data~\cite[Result 11]{Allen2025-canon}. Such experiments compare how readily architectures learn a skill from examples within a finite budget, rather than comparing similarly low scores after web pretraining. \textbf{They are also far cheaper}: one synthetic-playground run costs \mbox{$100$--$1000\times$} less than one real-life run at 1.3--8B on 1T tokens,%
\footnote{One synthetic-playground run takes about 10--128 H100 GPU-hours: its largest models (12 layers, width 768) train on 1--8 GPUs for less than 16 hours with unoptimized code, and smaller models take less time. One real-life run on 1T tokens takes about $7\times10^3$ H200 GPU-hours at 1.3B (on 128 GPUs) and $4\times10^4$ at 8B (on 512 GPUs; \appendixref{sec:setup:lr}), and an H200 GPU-hour delivers more training than an H100 GPU-hour.}
so the more than $10^6$ GPU-hours behind our 64 runs could instead pay for roughly $10^4$--$10^5$ such runs. Pretraining loss is no substitute for such tests: it is informative about language modeling (\figureref{fig:loss-linear} uses it), but it says little about in-context skills such as those in \namedref{Results}{res:6} and~\ref{res:7}. For example, at 8B, every linear model has lower WikiText perplexity than vanilla Llama (\tableref{tab:linear-main}), yet each one recalls information less reliably on 1-hop-$L$ with only 500 background tokens. Loss-based comparisons and scaling laws may therefore not identify the better starting point for reasoning.\looseness=-1

Standard benchmarks may also miss differences between optimizers. \citet{wen2025fantastic} find that the speedups of new optimizers over a well-tuned baseline shrink as models grow, attributing much of the earlier reported gains to poorly tuned baselines. Our architecture results suggest another possible reason for the smaller measured speedups. At larger scales, models trained with different well-tuned optimizers may all learn the skills measured by standard benchmarks equally well. Similar benchmark scores would then leave open whether the optimizers differ on other skills (\appendixref{app:future}).\looseness=-1

\section{Related Work}
\label{sec:related}

\parhead{Linear architectures}
Recurrent sequence models include GLA~\cite{yang2023gated}, Mamba and Mamba2~\cite{gu2023mamba,dao2024transformers}, GDN~\cite{yang2024gated}, RWKV~\cite{peng2023rwkv}, and HGRN2~\cite{qin2024hgrn2}. The empirical study closest to ours trains 8B Transformer, Mamba, Mamba2, and hybrid models on 1.1T and 3.5T tokens~\cite{waleffe2024empirical}. Our 8B/1T models generally score higher on the standard benchmarks that study reports. This suggests that our runs do not underestimate the linear models, including Mamba2. The comparison is only indicative because the studies differ in their data and evaluation versions. Beyond scale, our study differs in two ways:  it improves the GLA/GDN gates  and strictly aligns the linear models, and it compares standard benchmark scores with retrieval and reasoning (\namedref{Results}{res:3}--\ref{res:7}).

\parhead{Local mixing}
Short convolutions predate Canon: those in GLA, Mamba, and GDN follow H3~\cite{fu2022hungry}, and RWKV mixes each token with its predecessor before the projections in its time-mixing and channel-mixing blocks~\cite{peng2023rwkv}. Other designs place convolutions elsewhere: Primer convolves the query, key, and value projections~\cite{so2109primer}, Conformer~\cite{gulati2020conformer} and CvT~\cite{wu2021cvt} combine attention with convolution for speech and vision, and Multi-Token Attention convolves query--key maps~\cite{golovneva2025multi}. \empty Part 4.1 instead adds residual local connections at several positions in each layer, and this paper tests its recipes up to 8B/1T.

\parhead{Fair comparisons}
In optimizer research, re-tuning the baselines can substantially reduce reported speedups~\cite{wen2025fantastic}. We re-tune the architecture baselines as well. For comparisons among the linear models, we also match recurrent memory size, the parameter split between the recurrent layer and the MLP, and head geometry. Even a fair comparison can mislead if its tests miss the relevant skill: a study of 26 Transformer variants at 7--8B finds that short-context scores do not predict performance after context extension~\cite{bertsch2026cracks}, and we find that strong standard scores need not imply reliable retrieval even within short contexts.

\parhead{Copying and retrieval}
Prior work proves that models with a fixed-size state are limited in copying, and finds copying and phone-book retrieval weaknesses in such models~\cite{jelassi2024repeat}, including Mamba2 at 8B/3.5T~\cite{waleffe2024empirical}. These analyses trace the weakness to a state too small for the input, to too little input-dependent mixing~\cite{arora2024zoology}, or to limited memory that forces the model to decide what to keep before it sees the question~\cite{arora2024just}. Our test asks a model to recall one of a few stated birth years amid as few as 500 background tokens. This is a much simpler retrieval task, and the information to remember is far less than the recurrent state can hold. Even so, three strictly aligned linear-model families, with or without Canon, recall less reliably than Transformers of the same size (\resultref{res:6}), so state size alone does not explain the gap (Remark~\ref{rem:not-capacity}).

\parhead{Hybrids}
Hybrid models combine attention with linear layers: Jamba~\cite{lieber2024jamba} and Qwen3-Next~\cite{Qwen3next} alternate the two, whereas Falcon-H1~\cite{tiifalconh1} runs them in parallel. Such hybrids help retrieval: attention--Mamba2 hybrids improve 8B Phonebook retrieval~\cite{waleffe2024empirical}, consistent with attention handling retrieval from the input. In our exploratory hybrid runs, latent two-hop reasoning remained unreliable (\appendixref{sec:res6:hybrid}).

\parhead{Skill acquisition and post-training}
 Parts 2.2~\cite{YXLZ2024-gsm2} and 3.2~\cite{AL2023-knowledgeUB} show that  limited fine-tuning can fail to establish error correction or knowledge manipulation. Related findings include LIMA's superficial alignment hypothesis~\cite{zhou2023lima} and evidence that RLVR often improves access to reasoning paths the base model already has~\cite{yue2025limitsrlvr}, although prolonged RL can expand the base model's reasoning abilities, depending on the starting model and training duration~\cite{liu2025prorl}.  On the pretraining side, Part 3.1~\cite{AL2023-knowledge} shows how  knowledge augmentation (e.g., rephrasing the pretraining data) makes knowledge retrievable from the model's weights. The cited study also found that post-training could not repair failures to extract that knowledge in its setting. Nemotron-CC applies knowledge augmentation at scale, yet most of our models still cannot do latent two-hop reasoning, so knowledge-augmented data alone does not teach this reasoning skill. Curriculum learning~\cite{bengio2009curriculum} orders training examples from easy to hard. Our proposal also concerns when data is introduced, but for one specific skill: in-context reasoning examples should appear early in pretraining, because adding this skill later is costly, though not impossible (\sectionref{sec:res6:sft}).

\section{Conclusion}
\label{sec:conclusion}

With 64 from-scratch runs of Transformers and linear models at 1.3--8B parameters on 1--2T tokens of the same high-quality data, using tuned learning rates and strictly aligned linear models, we asked whether standard benchmarks see the architectural differences that matter. At this scale, they do see several such differences that academic-scale pretraining (1.3B/100B) all but hides. For Transformers, Canon layers consistently help: at 8B, LlamaCanon first reaches vanilla Llama's best final MMLU accuracy with about 30\% fewer pretraining tokens, a saving that full training curves reveal more reliably than final-checkpoint scores. For linear models, our low-rank output gates yield the stronger GLA5 and GDN2, and with Canon, the simple GLA5 matches GDN2 and outperforms Mamba2. A difference invisible at academic scale therefore need not be absent: the academic-scale experiment may simply be too small to detect it.

Both designs, Canon layers and our low-rank output gates, follow one principle, rooted in a central question of deep learning: \textbf{\emph{existence is cheap; learnability is everything}}. That an architecture can represent a computation does not mean pretraining will learn it, at least not efficiently. \resultref{res:5} suggests that the principle also holds for the whole recurrent layer: the extra machinery of GDN2 and Mamba2 appears to do much of what a short convolution does.

Beyond these differences, we also found that standard benchmarks miss some very basic in-context skills, even at 8B/1T. The first is retrieving information stated in the input: linear models match or exceed Transformers on standard benchmarks, yet with as few as 500 background tokens they do so less reliably than even a vanilla Transformer of the same size, suggesting that in hybrids it is attention that retrieves from the input (\resultref{res:6}). The second, latent two-hop reasoning, surprised us: we expected 8B parameters and 1T tokens of a strong corpus to be more than enough for arguably the simplest two-hop question, stated in a few short sentences, yet even the 8B Transformers reach only about 50\%, against 33.3\% for random guessing (\resultref{res:7}).

These deficits  challenge  architecture research based mainly on standard benchmark scores, and they also reversed the roles we had assigned to real-life pretraining and the synthetic playground of Part 4.1. We set out to treat real-life pretraining as ground truth for the playground, which had predicted several of the differences above in models about $80\times$ smaller (\tableref{tab:resonance}). For telling architectures apart, however, it is the playground that detects differences real-life pretraining misses, at \mbox{$100$--$1000\times$} lower cost per run. Architecture research should therefore test such skills directly, using synthetic playgrounds where real-life runs cannot yet reveal them. For practitioners,  one lesson is to include in-context reasoning examples in pretraining data from an early stage, rather than reserve them  for mid-training, let alone post-training.

Together with Parts 3.1 and 4.1, these results suggest a compact recipe for pretraining base models, at least up to 8B parameters: knowledge-augmented data, as in Nemotron-CC, plus Canon layers. With both, our 8B LlamaCanon matches Llama-3.1-8B at roughly 1/15 of its model-size$\times$token budget (\figureref{fig:scatter}).

\parhead{Reproducibility}
Recipes, code, 1-hop-$L$/2-hop-$L$ generators, results, and curves for all 64 runs are public, as are the 16 Transformer checkpoints, but not the linear-model weights (\appendixref{app:repro}).

 \parhead{Comparison with Part 4.1}
  For readers interested in a direct comparison of this work with Part 4.1, \tableref{tab:resonance} lists the playground's findings that this work tests.  Most of these gains were invisible at academic scale (1.3B/100B), and we measure them here. The retrieval gap for linear models persists at 8B/1T. Architectural differences in multi-hop reasoning remain invisible in real-life pretraining at both scales, because no model reliably learns even latent two-hop reasoning there.

\par\medskip\noindent
\begin{minipage}{\textwidth}
\captionsetup{type=table}
\centering
\footnotesize
\newcommand{\resYes}{\textcolor{mygreen}{\ding{51}}}
\newcommand{\resNo}{\textcolor{red!75!black}{\ding{55}}}
\newcommand{\resPart}{\textcolor{orange!85!black}{$\sim$}}
\newcommand{\src}[1]{{\color{gray}#1}}
\newcommand{\grp}[1]{\multicolumn{4}{@{}l}{\emph{#1}}}
\setlength{\tabcolsep}{4.5pt}
\renewcommand{\arraystretch}{1.1}
\begin{tabular}{@{}>{\raggedright\arraybackslash}p{0.285\textwidth}>{\centering\arraybackslash}p{0.1\textwidth}>{\raggedright\arraybackslash}p{0.235\textwidth}>{\raggedright\arraybackslash}p{0.30\textwidth}@{}}
\toprule
\textbf{Playground finding} & \textbf{Synthetic\newline $\leq$100M} & \textbf{Real-life, 1.3B/100B\newline \empty (Part 4.1, \src{R12})} & \textbf{Real-life, 1.3--8B/1--2T\newline (\bblue{this work})} \\
\midrule
\grp{Positive findings (gains)} \\
\addlinespace[2pt]
Canon improves Transformers & \resYes~\src{R2} & \resNo\ no clear gain on \texttt{lm-eval}, JRT, or BabiLong & \resYes\ in every matched comparison; $\sim$30\% fewer tokens (\namedref{Results}{res:1}--\ref{res:2}) \\
\addlinespace[1pt]
With Canon, quarter RoPE extrapolates better & \resYes~\src{R3} & \resYes\ beyond the training context & \resYes\ for the combined recipe, beyond the training context (\sectionref{sec:res6:canon-retrieval}) \\
\addlinespace[1pt]
Canon helps on top of built-in conv: GLA $>$ GDN $>$ Mamba2 & \resYes~\src{R6--8} & \resNo\ not visible & \resYes\ in almost every case we tested (\resultref{res:5}) \\
\addlinespace[1pt]
GLA with Canon matches GDN and beats Mamba2 & \resYes~\src{R9--10} & \resPart\ matches, but all within noise & \resYes\ GLA5 matches GDN2, beats Mamba2 up to 8B (\namedref{Results}{res:4}--\ref{res:5}) \\
\addlinespace[1pt]
A low-rank output gate strengthens GLA and GDN & \resYes~\src{not included} & \resNo\ not visible & \resYes\ in separate, unreported 1--3B ablations (\resultref{res:3}) \\
\midrule
\grp{Negative findings (deficits)} \\
\addlinespace[2pt]
Linear models retrieve poorly despite large recurrent states & \resYes~\src{R11} & \resYes\ gap at $\leq$500 tokens, but less convincing at this scale & \resYes\ persists at 8B/1T; \texttt{lm-eval} and NIAH-S miss it (\resultref{res:6}) \\
\addlinespace[1pt]
Linear models learn multi-hop more slowly than Transformers & \resYes~\src{R10--11} & \resNo\ all models fail two-hop & \resNo\ no model reliably beats random \textbf{even at} 8B/1T (\resultref{res:7}) \\
\bottomrule
\end{tabular}
\caption{\label{tab:resonance}\textbf{The synthetic playground's findings at academic scale (1.3B/100B) and at 1--2T tokens.} \resYes: observed; \resNo: not observed; \resPart: partly observed. \empty Gray: Part 4.1 result\ (``not included'': not reported).}
\end{minipage}\par\medskip

\appendix

\bigskip
\bigskip
\begin{center}
{\Huge
\textsc{Appendix}
}
\end{center}

This appendix contains the configurations, evaluation details, and complete results behind the main paper. \appendixref{app:config} expands on the experimental controls of \sectionref{sec:setup} and lists every training setting; \appendixref{app:eval} details the evaluations; \appendixref{app:design} gives the construction and cost of Canon and the low-rank gate; \appendixref{app:result-details} collects every reported score, the token-saving computation, and the linear models' training curves; \appendixref{app:discussion} reports our hybrid runs and future directions; and \appendixref{app:repro} describes the released code, results, and weights.

\section{Experimental Controls and Complete Configurations}
\label{app:config}

The first three subsections expand on the first three controls of \sectionref{sec:setup} (data, tuning, and alignment), and the last lists the remaining training settings.

\subsection{A Fixed, High-Quality Data Mix}
\label{sec:setup:data}

All 64 models train on the original high-quality split of Nemotron-CC v1 (Nemotron-CC-HQ)~\cite{su2025nemotroncc}. In Nvidia's same-budget 8B/1T runs, this data gives 60.1 on the 10-task average of \figureref{fig:scatter}, against 57.0 for DCLM~\cite{li2024datacomp} and 53.2 for FineWeb-Edu~\cite{penedo2024fineweb}. This 7-point spread exceeds Canon's gain on the same average, so the data must be held fixed when comparing architectures. One likely reason for this quality is that Nemotron-CC's rewriting applies \emph{knowledge augmentation\nocorr}, which \emph{we first discovered and proposed}\ in Part 3.1~\cite{AL2023-knowledge}. There,  a model often could not use a piece of knowledge, for example to answer a question about it, even after seeing it in exactly the same words 100 times. When the same knowledge was rewritten in varied forms, the model truly ``learned'' it and could extract it when needed.

\subsection{Tuning and Compute}
\label{sec:setup:lr}

\tableref{tab:weight-decay} lists the peak learning rates selected for the reported training runs after the broader search noted in \sectionref{sec:setup}, together with the weight decay used at each rate.

\par\medskip\noindent
\begin{minipage}{\textwidth}
\captionsetup{type=table}
\centering
\small
\setlength{\tabcolsep}{6pt}
\renewcommand{\arraystretch}{1.1}
\begin{tabular}{@{}l*{3}{>{\centering\arraybackslash}p{2.3cm}}@{}}
\toprule
\textbf{Setting} & $\mathrm{lr}=0.002$ & $\mathrm{lr}=0.003$ & $\mathrm{lr}=0.005$ \\
\midrule
Transformer 1.3B / 1T & $\mathrm{wd}=0.02$ & $\mathrm{wd}=0.0133$ & --- \\
Transformer 1.3B / 2T & --- & $\mathrm{wd}=0.0133$ & $\mathrm{wd}=0.008$ \\
Transformer 3B, 8B / 1T & $\mathrm{wd}=0.03$ & $\mathrm{wd}=0.02$ & --- \\
\addlinespace[2pt]
Linear 1.3B / 1T & $\mathrm{wd}=0.01$ & $\mathrm{wd}=0.00667$ & $\mathrm{wd}=0.004$ \\
Linear 3B / 1T & $\mathrm{wd}=0.005$ & $\mathrm{wd}=0.01$ & $\mathrm{wd}=0.006$ \\
Linear 8B / 1T & $\mathrm{wd}=0.01$ & $\mathrm{wd}=0.00667$ & --- \\
\bottomrule
\end{tabular}
\caption{\label{tab:weight-decay}\textbf{Learning rates and weight decay of all runs.} Each setting trains at two or three peak learning rates (lr), each with its own weight decay (wd); a dash means the rate is not used. Models with and without Canon use the same values, as do GLA5, GDN2, and Mamba2.}
\end{minipage}\par\medskip

\parhead{Baseline calibration}
It is hard to verify that a model is well tuned without retraining it, but two comparisons with public models indicate that our training setup is no worse than theirs. Nvidia trained a similar 8B Transformer on the same data with the same 1T-token budget; on the 10-task average of \figureref{fig:scatter}, it scores 60.1, whereas our vanilla 8B Llama scores $\approx$62.6. Likewise, our 8B LlamaCanon scores $\approx$63.6, slightly above Llama-3.1-8B (63.2), which was trained on about 15 times as many tokens.

\parhead{Compute}
Since one billion parameters trained on one trillion tokens ($1\,\text{B}\!\cdot\!\text{T}$) costs about 5200 H200 GPU-hours, a single 8B/1T run takes roughly $4\times10^4$ GPU-hours, which is why we sweep learning rates rather than repeat seeds (\sectionref{sec:setup}). The 64 reported runs together take about 1.2M GPU-hours.\empty  This total covers only these final runs, not the roughly 1M additional GPU-hours spent developing the strict-alignment protocol and the GLA5/GDN2 designs of \sectionref{sec:linear-upgrade}. 

\subsection{Strict Alignment of the Compared Models}
\label{sec:setup:align}

The per-block parameter budgets of strict alignment are given in \sectionref{sec:linear-upgrade:split} and \tableref{tab:config}. This subsection derives the recurrent-memory sizes, describes the aligned Mamba2, and explains how the models see the same data in the same order.

\parhead{Recurrent memory}
We count the scalar entries in each block's main recurrent KV or SSM state; short-convolution caches are counted separately. With the FLA defaults~\cite{yang2024fla} at $d=2048$, GLA uses 4 heads with total key width $d/2$ and total value width $d$, giving $4\times(256\times512)=256d$ entries. GDN uses 6 heads with 256-dimensional keys (total width $0.75d$) and values of width 512 per head ($2\times$ the key width), giving $6\times(256\times512)=384d$ entries. Mamba2 uses state size 128 on a $2d$-wide expanded input, giving $256d$ entries per layer. Its default uses two layers per block, so each block holds $512d$ entries. In our aligned models, GLA5 and GDN2 use $d/128$ heads with key and value head dimensions 128 and 256, and Mamba2 uses one layer per block with state size 128 and expansion 2, so \emph{all three hold $256d$ entries}.

\parhead{Aligned Mamba2}
The default Mamba2 has no MLP and instead uses two Mamba2 layers per block. Our aligned Mamba2 replaces the second Mamba2 layer with a gated MLP, so that each block holds one Mamba2 layer and one MLP, as in the other models. It keeps Mamba2's SSM update and short convolution unchanged, and all our Mamba2 conclusions concern this variant.

\parhead{Identical data order}
Within each Canon pair, and among GLA5, GDN2, and Mamba2 at a given learning rate, the models train on identical data batches in identical order. The one exception is the 1.3B GDN2 run with Canon at learning rate 0.005: it follows this order for about its first 50B tokens, after which its loss fluctuations match no other 1.3B run, so its data order most likely changed when the run was resumed. Its data still come from the same corpus, and \figureref{fig:loss-linear} does not use it. Among all other runs that share a data order, the loss fluctuations agree except briefly at loss spikes. Transformers and linear models use different batch geometries (\tableref{tab:config}). In the 1T runs that share a data order, each Transformer batch contains the same data as the corresponding 2--3 consecutive linear-model batches.

\par\medskip\noindent
\begin{minipage}{\textwidth}
\captionsetup{type=table}
\centering
\small
\setlength{\tabcolsep}{2.1pt}
\renewcommand{\arraystretch}{1.1}
\begin{tabular}{@{}llcccccccc@{}}
\toprule
& \textbf{arch.} & \textbf{tok.} & \textbf{layers} & $d$ & \textbf{heads} & \textbf{rec.\ mem.} & \textbf{layer+MLP} & \textbf{ctx}$\times$\textbf{batch} & \textbf{steps} \\
\midrule
\multirow{2}{*}{8B}
 & Llama & llama3 & 32 & 4096 & 32 (8 KV) & full & $2.5d^2{+}10.5d^2$ & $4096\times1536$ & 160k \\
 & GLA5/GDN2/Mamba2 & llama3 & 32 & 4096 & 32 & $256d$ & $6d^2{+}7d^2$ & $4096\times512$ & 480k \\
\midrule
\multirow{2}{*}{3B}
 & Llama & llama2 & 26 & 3072 & 24 & full & $4d^2{+}8d^2$ & $4096\times1536$ & 160k \\
 & GLA5/GDN2/Mamba2 & llama2 & 26 & 3072 & 24 & $256d$ & $6d^2{+}6d^2$ & $12288\times256$ & 320k \\
\midrule
\multirow{2}{*}{1.3B}
 & Llama & llama2 & 24 & 2048 & 32 & full & $4d^2{+}8d^2$ & $4096\times1024$ & 240k \\
 & GLA5/GDN2/Mamba2 & llama2 & 24 & 2048 & 16 & $256d$ & $6d^2{+}6d^2$ & $16384\times128$ & 480k \\
\bottomrule
\end{tabular}
\caption{\label{tab:config}\textbf{Model shapes and training budgets.} \emph{tok.} is the tokenizer; \emph{rec.\ mem.}, the number of entries in the main recurrent state per block (Transformers instead keep the full KV cache); \emph{layer+MLP}, the nominal parameters per block in the attention or recurrent layer and in the MLP; and \emph{ctx}$\times$\emph{batch}, the context length times the number of sequences per step. Rows show the 1T runs; the 1.3B Transformer's 2T run uses batch 2048 and 240k steps. Learning rates are in \tableref{tab:weight-decay}, and other settings in \appendixref{app:train-config}.}
\end{minipage}\par\medskip

\subsection{Other Training Settings}
\label{app:train-config}

\tableref{tab:config} gives the shapes, batches, and steps, and \tableref{tab:weight-decay} the learning rates; the released recipes contain everything else. All models use AdamW ($\beta_1{=}0.9$, $\beta_2{=}0.95$, $\epsilon{=}10^{-8}$), gradient clipping at 1.0, a cosine schedule down to $1\%$ of peak, bf16 with FSDP full sharding, and a fixed seed.

\parhead{Transformers} At 1.3B, 3B, and 8B, the gated-MLP widths $d_{\mathrm{ffn}}$ are 5632, 8192, and 14336, respectively. The corresponding warmup periods are 8k, 4k, and 3k steps. The 8B model matches Llama-3.1-8B's shape, including its 8 KV heads, and uses its llama3 tokenizer (128{,}256 entries); our 8B RoPE base is $\theta_{\mathrm{RoPE}}{=}10^5$. The 1.3B and 3B models use llama2 (32k), since at $d=2048$ the llama3 vocabulary would add about 394M embedding parameters and distort the size comparison.

\begin{remark}[on QK-norm and $z$-loss]
QK-norm~\cite{henry2020query} and $z$-loss~\cite{chowdhery2023palm} are meant to stabilize training. Our 8B runs, the largest here, use neither and train stably, so neither is needed even at 8B. Our 1.3B and 3B recipes happen to keep both, and we did not have the compute to repeat these runs without them. Both models in each Canon pair make the same choice, so Canon's gain is measured within each recipe.
\end{remark}

\parhead{Linear models} At 1.3B, 3B, and 8B, the gated-MLP widths $d_{\mathrm{ffn}}$ are 4096, 6144, and 9472, respectively. The corresponding warmup periods are 12k, 8k, and 9k steps. Each run uses the same 1T-token budget as its Transformer counterpart ($\text{steps}\times\text{batch}\times\text{context}\approx1.007\times10^{12}$).

\section{Evaluation Details}
\label{app:eval}
\label{sec:setup:eval}

The five evaluation groups below are reported separately, not combined into one score.

\begin{itemize}\setlength{\emergencystretch}{1em}
\item For the \bblue{\texttt{lm-eval}} score~\cite{eval-harness}, we average 13 accuracy metrics (LAMBADA~\cite{paperno2016lambada}, PIQA~\cite{bisk2020piqa}, HellaSwag~\cite{zellers2019hellaswag}, WinoGrande~\cite{sakaguchi2019winogrande}, ARC-easy and ARC-challenge~\cite{clark2018think}, SIQA~\cite{sap2019social}, BoolQ~\cite{clark2019boolq}, COPA, OpenBookQA, RACE, CommonsenseQA, MMLU~\cite{hendrycks2021measuring}). We report WikiText and LAMBADA perplexity separately.

\item \bblue{JRT} \cite{arora2024just}: 12 generative/retrieval metrics (FDA, SWDE, SQuAD~\cite{rajpurkar-etal-2018-know}, TriviaQA~\cite{joshi-etal-2017-triviaqa}, NQ~\cite{kwiatkowski-etal-2019-natural}, DROP~\cite{dua-etal-2019-drop}, each in a single-pass and a read-twice variant). The read-twice variant repeats the prompt, letting a recurrent model revisit the document after seeing the question.

\item \bblue{BabiLong} \cite{kuratov2024babilong}: bAbi~\cite{weston2015towards} tasks qa1--qa5 embedded in junk text at 0k/1k/2k/4k, 20 metrics.

\item \bblue{NIAH-S} \cite{hsieh2024ruler}: single-needle retrieval of a ``special magic number'' for a key word, hidden in repeated filler sentences or essays; the value is a random seven-digit number in two variants and a UUID in the third.

\item \bblue{1-hop-$L$ / 2-hop-$L$}:  our  one- and two-hop retrieval tasks (Sections~\ref{sec:res6:task} and~\ref{sec:res7}), with background length $L=0$, $0.5$k, and $1$k--$6$k. They complement broader suites such as RULER~\cite{hsieh2024ruler} and LongBench~\cite{bai2024longbench} by keeping the statements simple and varying only the unrelated text and the number of hops.  The generators are public. \end{itemize}

\parhead{Score conventions} The \texttt{lm-eval} average weights its 13 accuracy metrics equally. \figureref{fig:scatter} instead uses the 10 tasks that Nvidia also reports~\cite{su2025nemotroncc}, omitting LAMBADA, BoolQ, and COPA, and its points are individual runs rather than best-of-rate summaries. We re-evaluate the public models in our harness; Nvidia's checkpoints are unavailable, so their points use the published scores.

\parhead{Two evaluation pipelines} Final scores use the harness's \texttt{HFLM} implementation in mixed-precision bf16 on the full evaluation sets; training curves use Lingua's integrated pure-bf16 evaluation every 2k--6k steps. The two can differ by a few tenths of a point on the same checkpoint, so each comparison uses only one of them.

\parhead{Metric conventions} All \texttt{lm-eval} tasks, MMLU included, are evaluated zero-shot. HellaSwag, ARC-challenge, and OpenBookQA use \texttt{acc\_norm}; other standard tasks use \texttt{acc}. WikiText uses \texttt{word\_perplexity}, and LAMBADA uses last-word \texttt{perplexity}. JRT uses \texttt{contains} rather than exact match or F1. BabiLong and NIAH-S retain their original metrics.

\parhead{1-hop / 2-hop prompts}\phantomsection\label{app:eval:prompts}  Names come from Part 4.1's \textsc{Capo} name lists,  distinct years are drawn uniformly from 1950--2009, and background text comes from a held-out Wikipedia validation split. In 1-hop, five people each receive a different year, stated as \detext{``[Name] was born in the year of [Year].''}, and the query is \detext{``Answer me: [Name] was born in the year of ''}. In 2-hop, we give three people birth years and pair them one-to-one and at random with three new people. We state each pairing as \detext{``[Name$_j$] was born in the same year as [Name$_i$].''} The query uses the same completion format. The statements keep their order when interleaved with the background, with the birth-year statements first in 2-hop. The query asks about the first of the five people in 1-hop and the first newly introduced person in 2-hop. In 1-hop, the last person's birth year is the closest of the stated years to the query. Asking about that person could let the model copy the year without matching the name, so we ask about the first person. The 2-hop prompt starts with a short instruction, \detext{``You will be asked questions about people's birth years, and the birth year descriptions are hidden in some random text. Some people's birth years are directly given, while others are given in the form that `name1' was born in the same year as `name2'.''} It then places the interleaved statements and background inside \texttt{<context>} tags and appends the query after the closing tag, using the same completion format as in 1-hop. The reported results for both tasks use zero-shot prompts. For 2-hop, we also tried prompts with three worked examples and prompts without the instruction; neither changed accuracy noticeably. We kept the zero-shot prompt with the instruction because it is shorter than the worked-example prompt, so failures are harder to attribute to input length.

\parhead{Scoring} Each task uses 1,000 examples at each background length. The model generates greedily for at most 10 new tokens, and the answer counts as correct if the decoded continuation contains the correct year. Prompts are not truncated, so at $L=4$k the statements, query, and wrapper can push the full prompt past a 4096-token training window. Random guessing among the stated years scores 20\% on 1-hop-$L$ (five years) and 33.3\% on 2-hop-$L$ (three years).

\section{Architecture Details: Canon Layers and Low-Rank Gates}
\label{app:design}

\subsection{Canon Construction, Parameters, and RoPE}
\label{app:canon-details}

This subsection summarizes the Canon construction so that our models can be understood  without reading Part 4.1~\cite{Allen2025-canon}, which develops  the design and its synthetic experiments in full.

A simple example shows what Canon does in the first layers. Consider predicting the final token in \texttt{[A] [B] $\ldots$ [A] [?]}. The model should associate the earlier \texttt{[A]} with its following \texttt{[B]}, but causal attention cannot let that first \texttt{[A]} look ahead. One layer can instead copy the \texttt{[A]} representation into the neighboring \texttt{[B]} position, and a later layer can match the final \texttt{[A]} to that position and retrieve \texttt{[B]}. Canon provides this first, local step directly, so attention need not be spent on it. In recurrent models, it similarly spares the recurrent state from carrying such nearby information.\looseness=-1

This example is only illustrative, and Canon matters in later layers as well~\cite[Result 2.1]{Allen2025-canon}.  In Part 4.1, Canon in the first block alone  recovered only a small fraction of its gain in multi-hop reasoning. Most of the gain instead comes from later layers: there, Canon acts like a residual connection along the token axis, which lets a model learn deep tasks hierarchically (first 1-hop, then 2-hop, and so on) much faster.

The operation in \eqref{eqn:canon} is a depthwise causal convolution with kernel size 4: each channel has its own weights, shared across positions, and missing left context is zero-padded. Its name recalls a musical canon, in which a melody returns at fixed delays.

The width $m$ of a Canon layer depends on its insertion point (\sectionref{sec:canon-recap}). In a Transformer block, Canon-A and Canon-C each have width $d$ and act on the RMSNorm outputs before attention and the MLP, respectively. Canon-B acts after the Q/K/V projections. Its width is $(n_{\mathrm{heads}}{+}2n_{\mathrm{kv}})d_{\mathrm{head}}$, where $n_{\mathrm{heads}}$ counts query heads, $n_{\mathrm{kv}}$ counts key/value heads, and $d_{\mathrm{head}}$ is the head dimension. This gives $3d$ with one key/value head per query head and $1.5d$ with grouped KV in our 8B model. Canon-D covers both branches of the gated MLP, each of width $d_{\mathrm{ffn}}$, so its width is $2d_{\mathrm{ffn}}$. In linear models, position B holds the built-in convolution (Canon-b),  which Part 4.1 found especially important  for GLA and Mamba2 and less so for GDN.

\parhead{Parameter count}
A Canon layer of width $m$ has $4m$ weights and no bias, so Canon-ABCD adds $4(2d+(n_{\mathrm{heads}}{+}2n_{\mathrm{kv}})d_{\mathrm{head}}+2d_{\mathrm{ffn}})$ per block: 2.06M, 3.30M, and 5.51M parameters in total at 1.3B, 3B, and 8B (models of 1.364B, 3.141B, and 8.030B with untied embeddings), or \bblue{0.151\%, 0.105\%, and 0.069\%}. The fraction falls with scale because Canon costs $O(d)$ per block while the main projections cost $O(d^2)$.

\parhead{Canon and RoPE}
\empty Part 4.1~\cite[Result 3]{Allen2025-canon} found that with Canon, applying RoPE to only 1/4 of the dimensions works better than either full RoPE or no RoPE, and also helps length generalization. This is because much of the local work for which attention relies on RoPE, such as locating nearby tokens, can now be done by Canon. Canon-ABCD already improves Transformers with full RoPE~\cite[Result 2]{Allen2025-canon}; the quarter RoPE is a refinement that Canon makes possible. LlamaCanon therefore uses this quarter RoPE with Canon-ABCD, while vanilla Llama uses full RoPE; GPT-NeoX-20B likewise applies rotary embeddings to one quarter of each head's dimensions~\cite{gpt-neox-20b}. Our runs at scale do not separate Canon from the quarter RoPE, so the Transformer gains reported here belong to the Canon recipe as a whole.

\par\medskip\noindent
\begin{minipage}{\textwidth}
\captionsetup{type=figure}
\centering
{\includegraphics[page=2,trim={3mm 93mm 2mm 0mm},clip,width=0.99\textwidth]{plots2}}
\caption{\label{fig:canon-illustrate}\textbf{Canon layers and their four insertion points}, reproduced from Part 4.1~\cite{Allen2025-canon}.}
\end{minipage}\par\medskip

\subsection{Step-Time Cost of Canon}
\label{app:canon-cost}

Canon adds only 0.07--0.15\% to the parameters but more work per step.  Part 4.1 measured forward/backward overheads  of 12.4\%/14.1\% for a 1.3B Llama with Canon-ABCD on H100s, using a naive implementation that relied on \texttt{torch.compile} defaults. With this $\sim$13\% step-time overhead, the $\sim$30\% token saving of \resultref{res:2} amounts to about \bblue{20\% less training time} ($0.7\times1.13\approx0.79$).

\parhead{Fused implementations}
We have since fused each Canon layer with a neighboring operation and compared the fused implementation with the best kernels of Transformer Engine~2.20.2~\cite{nvidia2026transformerengine} in Megatron~\cite{shoeybi2019megatron}. These fused kernels are preliminary and not yet fully optimized, so the costs below may still fall. The test model is a 4B dense Transformer with 32 layers, $d=3072$, and three sliding-window attention layers (window 2048) per full-attention layer. We trained it with an 8192-token context in NVFP4 mixed precision, using NVFP4 for linear-layer matrix multiplications and BF16 for other operations. We used a per-GPU batch size of 6 on GB300 and 4 on GB200 because GB200 GPUs have less memory. We measured end-to-end step time in each comparison.\looseness=-1
\begin{itemize}
\item Canon-B, fused in Triton with the QKV split and RoPE, is 0.8--2.0\% \emph{faster} than Megatron's standard path of QKV split plus Transformer Engine's fused RoPE.
\item Canon-D, fused with SwiGLU (with a CUDA backward kernel), costs 0.7--0.9\% more than Transformer Engine's SwiGLU.
\item Canon-A and Canon-C, each fused in Triton with the RMSNorm before it, together cost 3.2--3.6\% more than Transformer Engine's RMSNorm, which runs on cuDNN~9.27.
\item All four together, Canon-ABCD, cost 2.1\% on GB300 and 3.3\% on GB200.
\end{itemize}
In the 4B benchmark, fused Canon-ABCD adds 2.1--3.3\% to step time, below the 13\% overhead assumed above for the naive implementation. If a similar overhead holds for the 8B runs, their token saving in \resultref{res:2} would translate to \bblue{more than 20\% less training time} at the same number of tokens per step.

\subsection{Low-Rank Gate Accounting and Initialization}
\label{app:gate-details}

GLA5 and GDN2 use the gate of \eqref{eqn:gate-new} with rank $r=16$ and total value dimension $d_v=2d$. The nominal per-block budgets of \sectionref{sec:linear-upgrade:split} omit the small parameter counts of the rank-16 output gates in GLA5 and GDN2, as well as the convolutions and normalization. At $d_v=2d$, the full-rank gate discussed in \sectionref{sec:linear-upgrade:gate} would take $2d^2$ parameters per block. The rank constraint applies to $G_1G_2$ before the elementwise SiLU, so the activated gate itself need not be low-rank.

\parhead{Initialization}
For independent, zero-mean draws, using the original entrywise standard deviation $\sigma$ of $G$ for both $G_1$ and $G_2$ would give each entry of $G_1G_2$ variance $r\sigma^4$ instead of $\sigma^2$. We therefore draw both factors with
\begin{equation}\label{eqn:gate-init}
\sigma_{\mathrm{low}} \;=\; \sigma^{1/2} \cdot r^{-1/4},
\end{equation}
so that $r\sigma_{\mathrm{low}}^4=\sigma^2$, and initialize $b$ to zero. For any fixed input $x$, $G_1G_2x$ and $Gx$ then have the same expected squared norm, $d_v\sigma^2\|x\|_2^2$, although their full distributions differ.

\section{Complete Results}
\label{app:result-details}

This appendix collects every reported score, together with the token-saving computation and the linear models' training curves.

\subsection{Per-Task Scores}
\label{app:config-summaries}
\label{app:all-lr}
\label{app:one-hop-results}
\figureref{fig:trans-table} and \figureref{fig:linear-table} show the best score across learning rates for each reported metric and configuration. \figureref{fig:trans-table-all} and \figureref{fig:linear-table-all} show each run separately, with all its task scores together. All four figures show retrieval and latent two-hop reasoning scores, including at lengths beyond the training context.

\parhead{Retrieval and two-hop columns}
The 1-hop-$L$ and 2-hop-$L$ columns of \figureref{fig:linear-table} give the per-configuration scores behind \figureref{fig:hop-length}. To see how Canon affects retrieval for GLA5 or GDN2, compare its Canon-AbCD row with its \texttt{conv} row (Canon-b only) at the same size and background length in the 1-hop-$L$ columns. Lengths of 5k--6k exceed the training window only for the Transformers and the 8B linear models (\sectionref{sec:setup}). Beyond 4k, LlamaCanon extrapolates better than vanilla Llama (\sectionref{sec:res6:canon-retrieval}), likely because of its quarter RoPE as well as local mixing,  as Part 4.1's ablations suggest.  The linear models, which have no explicit positional encoding, lose accuracy more gradually than the Transformers as the background grows, and the best of them beat vanilla Llama at 5k and 6k. On 2-hop-$L$, many configurations fall below random guessing at longer backgrounds. We treat each gap between a Canon model and its baseline as specific to the model and background length being compared.

\par\medskip\noindent
\begin{minipage}{\textwidth}
\captionsetup{type=figure}
\centering
\setlength{\imgwidthBase}{0.99\textwidth}
\includegraphics[width=\imgwidthBase]{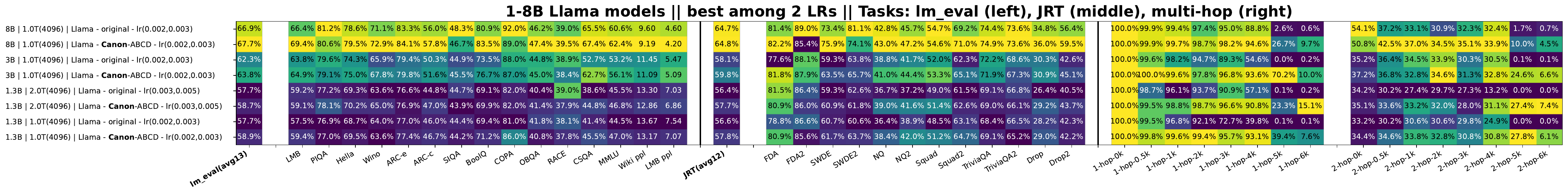}
\\ \smallskip
\includegraphics[width=\imgwidthBase]{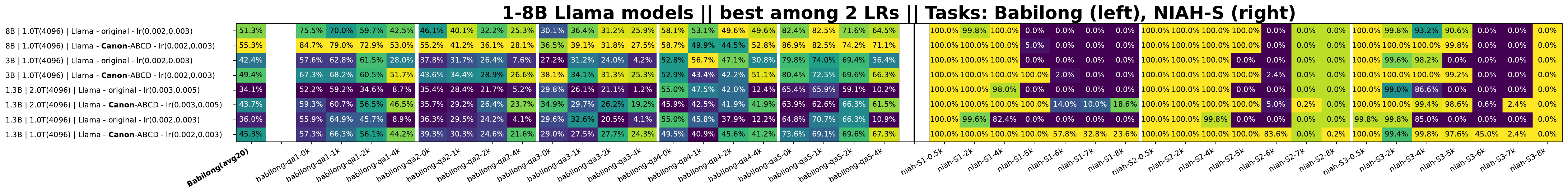}
\caption{\label{fig:trans-table}\textbf{Per-task scores of the eight Transformer configurations.} Each metric takes the better of two learning rates, and averages are computed as in \tableref{tab:trans-main}. The top table covers \texttt{lm-eval}, JRT, 1-hop-$L$, and 2-hop-$L$, and the bottom table covers BabiLong and NIAH-S. Each vanilla row is followed by its Canon version. For every run separately, see \figureref{fig:trans-table-all}.}
\end{minipage}\par\medskip

\par\medskip\noindent
\begin{minipage}{\textwidth}
\captionsetup{type=figure}
\centering
\setlength{\imgwidthBase}{0.99\textwidth}
\includegraphics[width=\imgwidthBase]{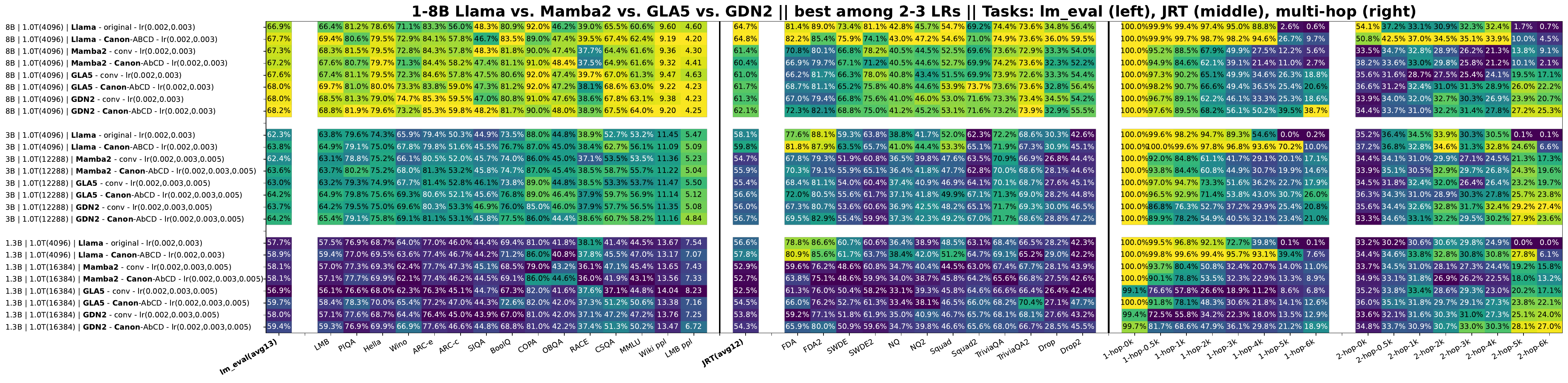}
\\ \smallskip
\includegraphics[width=\imgwidthBase]{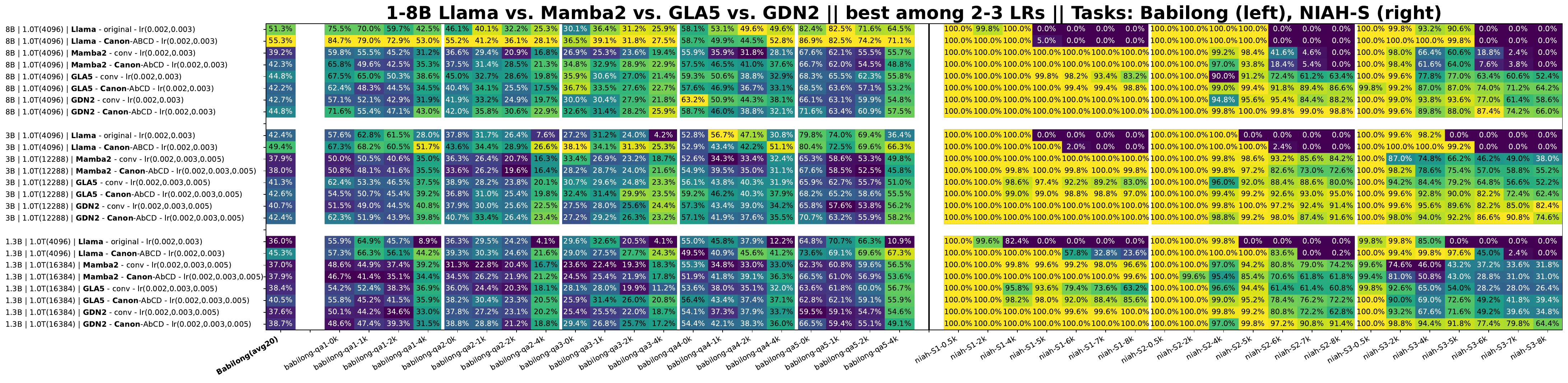}
\caption{\label{fig:linear-table}\textbf{Per-task scores of all 24 configurations at 1T.} Each metric takes the best of 2--3 learning rates, and averages are computed as in \tableref{tab:linear-main}. Each size block lists Llama, Mamba2, GLA5, and GDN2, each followed by its Canon version. The top table covers \texttt{lm-eval}, JRT, 1-hop-$L$, and 2-hop-$L$, and the bottom table covers BabiLong and NIAH-S. For every run separately, see \figureref{fig:linear-table-all}.}
\end{minipage}\par\medskip

\par\medskip\noindent
\begin{minipage}{\textwidth}
\captionsetup{type=figure}
\centering
\setlength{\imgwidthBase}{0.99\textwidth}
\includegraphics[width=\imgwidthBase]{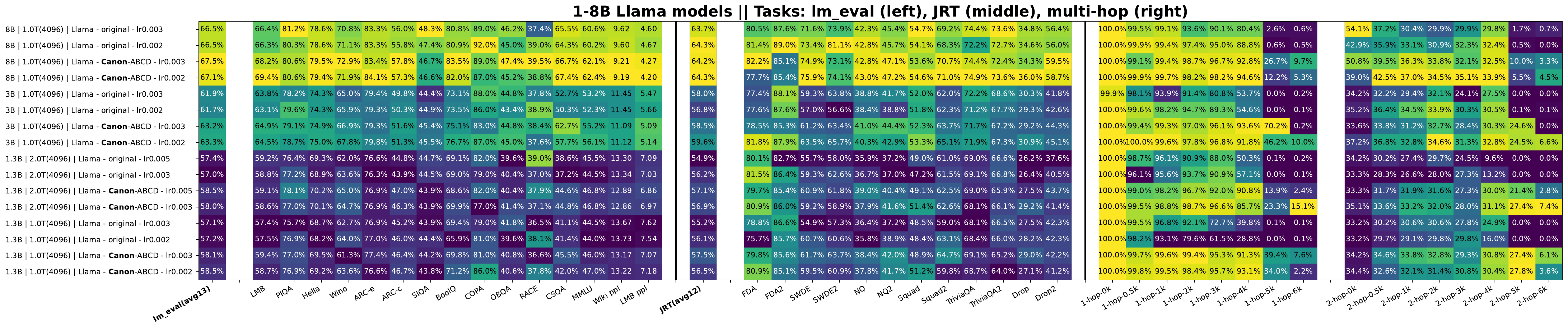}
\\ \smallskip
\includegraphics[width=\imgwidthBase]{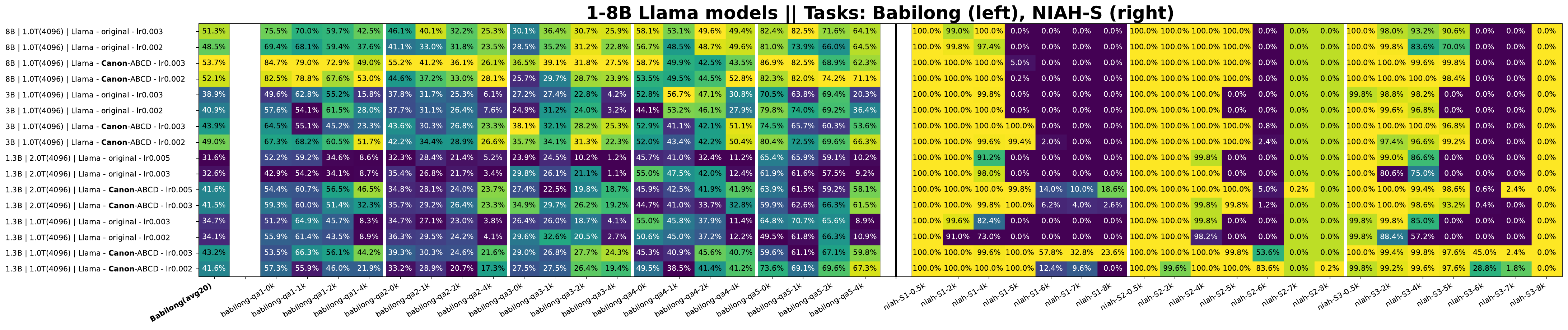}
\caption{\label{fig:trans-table-all}\textbf{All 16 Transformer runs, each learning rate shown separately.} Same layout as \figureref{fig:trans-table}, which instead takes each metric's best learning rate.}
\end{minipage}\par\medskip

\par\medskip\noindent
\begin{minipage}{\textwidth}
\captionsetup{type=figure}
\centering
\setlength{\imgwidthBase}{0.99\textwidth}
\includegraphics[width=\imgwidthBase]{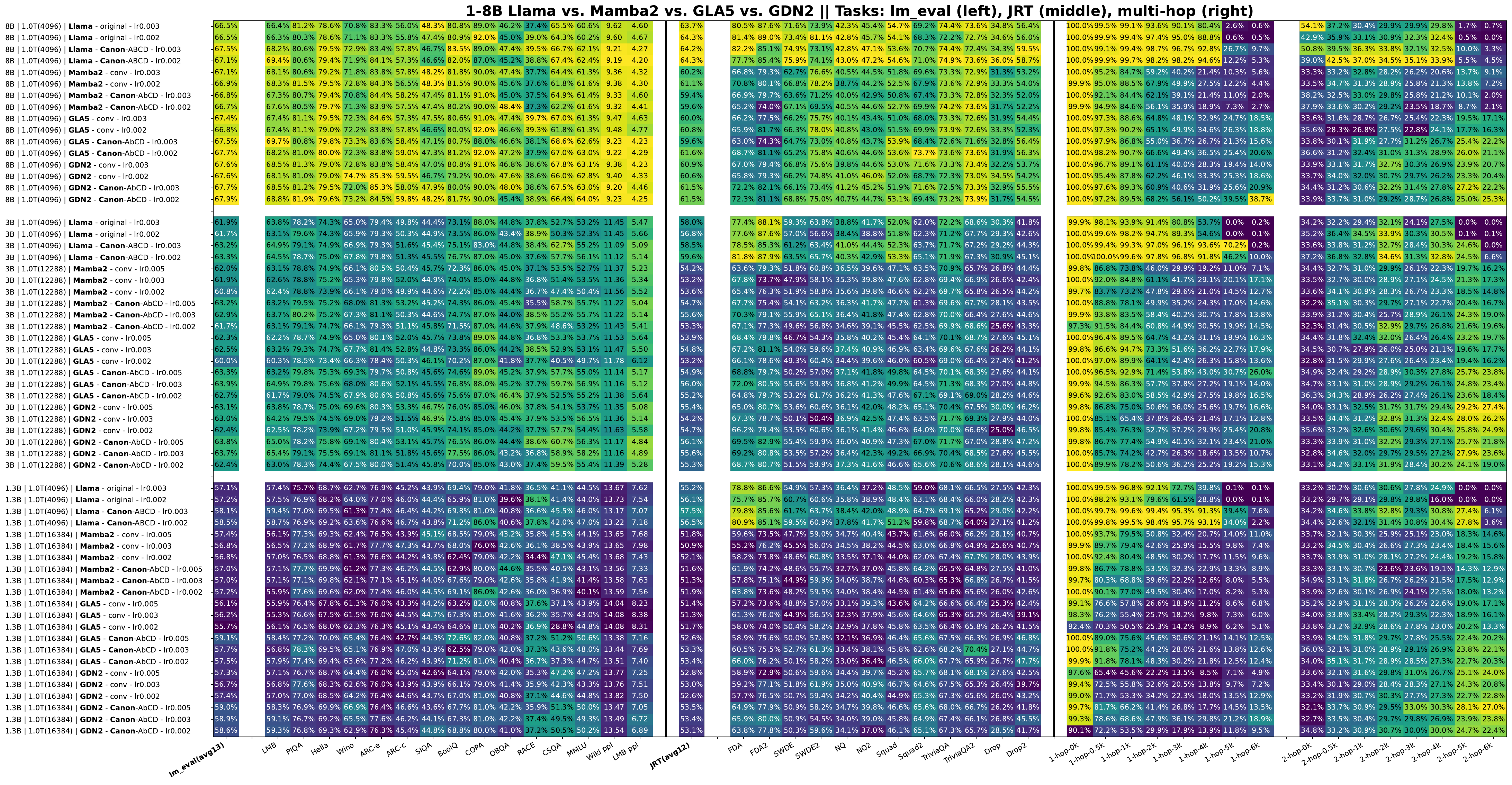}
\\ \smallskip
\includegraphics[width=\imgwidthBase]{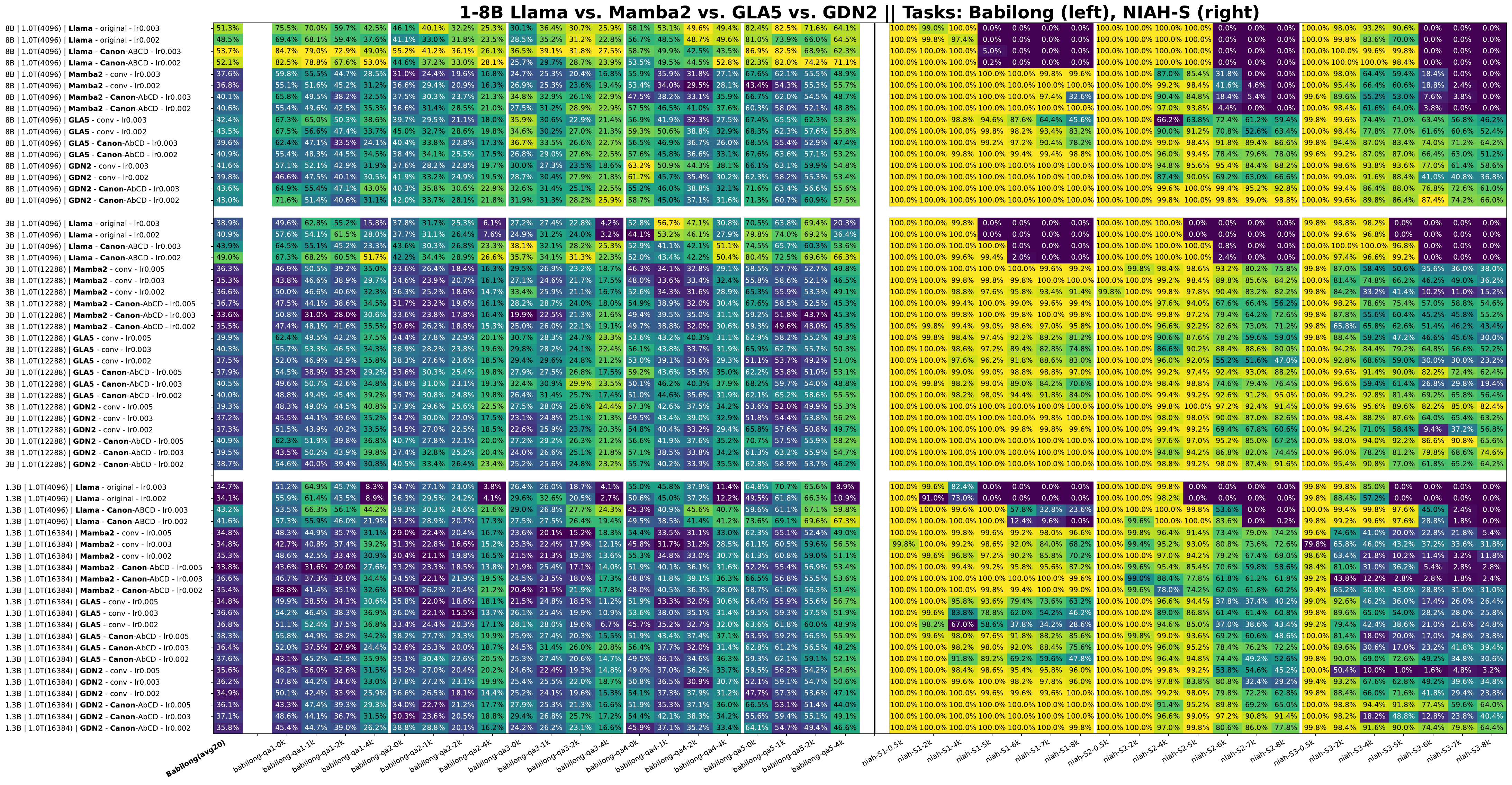}
\caption{\label{fig:linear-table-all}\textbf{All 48 linear runs and 12 Transformer reference runs at 1T, each learning rate shown separately.} Same layout as \figureref{fig:linear-table}, which instead takes each metric's best learning rate.}
\end{minipage}\par\medskip

\subsection{Token-Saving Computation}
\label{app:token-details}

At each checkpoint, we take the highest MMLU score across learning rates separately for the model with Canon and its baseline, as in \sectionref{sec:res2}. We then interpolate linearly between checkpoints to find the first token count $t_C$ at which the model with Canon reaches the final accuracy of its baseline without Canon. If $T$ is the baseline's final token count, the saving is $1-t_C/T$. For LlamaCanon, this gives 27.7\%, 32.0\%, and 25.5\% at 8B, 3B, and 1.3B on 1T tokens, and 36.7\% at 1.3B on 2T tokens. At 8B, smoothing each Canon curve with a centered moving average of 1--15 checkpoints gives 25.5--28.1\%, and moving the target by $\pm1$ point gives 26.5--33.9\%. The 1.3B curves are noisier, so their estimates move more: 20.6--46.0\% at 1T under the target shift, and 21.1--36.7\% at 2T under smoothing.

\subsection{Training Curves of the Linear Models}
\label{app:linear-curves}

\figureref{fig:curve-linear} complements the training-loss comparison in \figureref{fig:loss-linear} with MMLU curves for GLA5 and GDN2, using the run with the highest final MMLU score for each configuration. The token savings in \resultref{res:4} use the best score across learning rates at each checkpoint, following \appendixref{app:token-details}.

\parhead{Training loss}
At matched learning rates, Canon lowers the training loss most for GLA5 and least for Mamba2 in all eight size$\times$rate settings, whether the loss is averaged over the last 25B, 50B, 100B, 200B, or 500B tokens (\resultref{res:5}).

\par\medskip\noindent
\begin{minipage}{\textwidth}
\captionsetup{type=figure}
\centering
\includegraphics[width=0.8\textwidth]{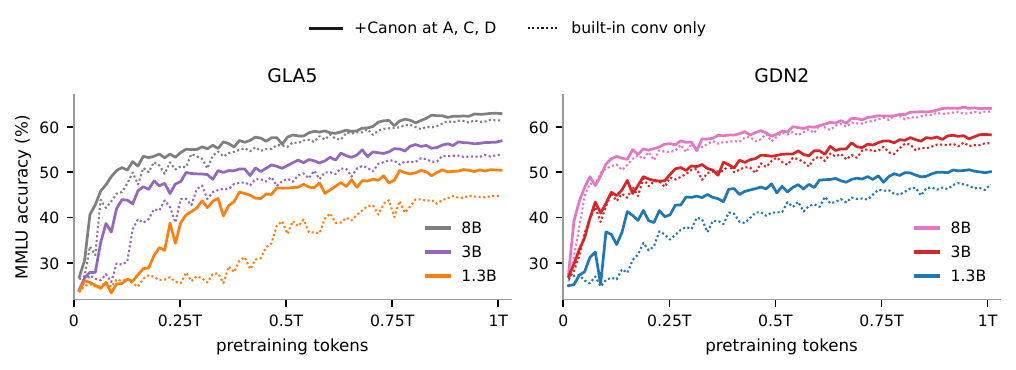}
\caption{\label{fig:curve-linear}\textbf{With Canon, GLA5 and GDN2 follow similar MMLU training curves.} GLA5 (left) and GDN2 (right) at 1.3B, 3B, and 8B, each run at the learning rate with the best final MMLU. Solid curves add Canon at A, C, and D, and dotted curves use only the built-in convolution. The token savings of \resultref{res:4} are computed as in \resultref{res:2}, from the better learning rate at each checkpoint.}
\end{minipage}\par\medskip

\section{Hybrid Runs and Future Directions}
\label{app:discussion}
\label{sec:res6:hybrid}
\label{app:future}

\parhead{Hybrid runs}
Beyond the 64-run suite, we also trained hybrids in a similar setting: with interleaved attention and linear layers, with parallel attention and recurrent branches, and with learned gates, across several attention fractions. None of these hybrid runs showed reliable latent two-hop reasoning. We report only that outcome here. Future hybrid experiments could directly test the division of labor proposed in \sectionref{sec:hybrid-labor}\ by varying the attention fraction and measuring 1-hop-$L$ accuracy at long background lengths.

\parhead{Future directions}
\begin{itemize}
\item \textsc{Retrieval inside linear layers.}
\resultref{res:6} motivates improving how linear layers store information stated in their input and later retrieve it. Our accuracy measurements alone cannot tell whether a linear model fails when storing the information in its recurrent state or when reading it back. Hybrids would lose one concrete reason to keep attention layers only if a future linear model could retrieve as reliably as a Transformer of the same size on 1-hop-$L$. The background text would have to be about as long as the linear model's training window (e.g., 4k tokens, far shorter than what is usually called long context).
\item \textsc{Optimizer comparisons beyond standard scores.}
Optimizers could be compared like our architectures, under matched tuning, on both standard benchmarks and reasoning tasks with explicit training examples, to test whether the shrinking gains reported by \citet{wen2025fantastic} depend on which skills are measured (\sectionref{sec:res7:lessons}).
\item \textsc{Cheaper Canon.}
With our preliminary fused kernels, Canon-B and Canon-D add almost no step time. Canon-A and Canon-C account for most of the added time per step (\appendixref{app:canon-cost}). A cleaner way to provide the same local mixing, such as the cross-layer variant $h'_{\ell+1}=h_{\ell+1}+\mathrm{Canon}(h_\ell)$ or a different efficient form of local mixing (\sectionref{sec:canon-recap}), may remove most of this remaining cost and turn more of the roughly 30\% token saving into saved training time.
\end{itemize}

\section{Reproducibility}
\label{app:repro}

\def\ShowRepro{1}\ifdefined\ShowRepro
\parhead{Code and recipes} The original release is at \href{https://github.com/facebookresearch/PhysicsLM4}{\texttt{facebookresearch/\allowbreak PhysicsLM4}}, with continuing maintenance at \href{https://github.com/zhuzeyuan/PhysicsLM4}{\texttt{zhuzeyuan/\allowbreak PhysicsLM4}}. The \texttt{canon\_\allowbreak llama\_\allowbreak recipes/} and \texttt{canon\_\allowbreak linear\_\allowbreak recipes/} directories specify the reported training runs. Training uses a modified Meta Lingua~\cite{meta2024lingua}, licensed under BSD-3-Clause, while our own contributions use Apache 2.0.

\parhead{Reported results} The public \href{https://github.com/facebookresearch/PhysicsLM4/blob/75806965208c55e9f7f2e2363d3793d2da75d2e4/canon_llama_results/README.md}{Transformer results} and \href{https://github.com/facebookresearch/PhysicsLM4/blob/75806965208c55e9f7f2e2363d3793d2da75d2e4/canon_linear_results/README.md}{linear results} directories contain complete result tables, evaluation outputs, and interactive training curves; the linear-model directory also contains the training-loss curves behind \figureref{fig:loss-linear}. Download the linked HTML files and open them locally to inspect the curves.

\parhead{Released weights} The 16 Transformer checkpoints are hosted on Hugging Face and linked from \texttt{canon\_llama\_recipes/}. Only these Transformer weights were cleared for release. The 48 linear-model runs have public recipes and evaluation outputs but not public weights, so reproducing these models requires retraining.

\fi

\small
\setlength{\bibsep}{3pt}

\bibliographystyle{plainnat}

\bibliography{canon}

\end{document}